\documentclass[10pt,twocolumn,letterpaper]{article}
\usepackage{cvpr}

\usepackage{graphicx}
\usepackage{amsmath}
\usepackage{amssymb}
\usepackage{algorithmicx}
\usepackage{algorithm}
\usepackage{algpseudocode}%
\usepackage{booktabs}
\usepackage{multirow}
\usepackage{comment}
\usepackage{pifont}%
\usepackage{enumerate}
\usepackage{enumitem}
\usepackage[accsupp]{axessibility}
\usepackage{tabularx}
\usepackage{makecell}
\usepackage{soul}
 
\usepackage{ulem}
\usepackage{color}
\usepackage{commath}

\definecolor{bronze}{rgb}{1,1,0.6}
\definecolor{silve}{rgb}{0.969,0.796,0.600}
\definecolor{gold}{rgb}{0.941,0.592,0.600}
\definecolor{purple}{rgb}{0.913,0.753,0.941}
\newcommand{\gold}[1]{\colorbox{gold}{{#1}}}
\newcommand{\silve}[1]{\colorbox{silve}{{#1}}}
\newcommand{\bronze}[1]{\colorbox{bronze}{{#1}}}

\usepackage{xspace}

\newcommand\nth{\textsuperscript{th}\xspace}

\definecolor{cvprblue}{rgb}{0.21,0.49,0.74}
\usepackage[pagebackref,breaklinks,colorlinks,citecolor=cvprblue]{hyperref}

\begin{document}

\title{Real-time Free-view Human Rendering from Sparse-view RGB Videos using \\ Double Unprojected Textures}

\author{ 
    Guoxing Sun\textsuperscript{1},
    Rishabh Dabral\textsuperscript{1,2},
    Heming Zhu\textsuperscript{1},
    Pascal Fua\textsuperscript{3},
    Christian Theobalt\textsuperscript{1,2},
    Marc Habermann\textsuperscript{1,2}  
    \\
\normalsize{\textsuperscript{1}Max Planck Institute for Informatics, Saarland Informatics Campus}
    \quad
    \normalsize{\textsuperscript{2}VIA Research Center}
    \quad
    \normalsize{\textsuperscript{3}EPFL}\\
    \normalsize{\{gsun, rdabral, hezhu, theobalt, mhaberma\}@mpi-inf.mpg.de \quad pascal.fua@epfl.ch}\\
    \normalsize{\url{https://vcai.mpi-inf.mpg.de/projects/DUT/}}
}

\twocolumn[{%
\renewcommand\twocolumn[1][]{#1}%
\maketitle
\begin{center}

     \vspace{-25pt}
     \includegraphics[width=1.0\linewidth]{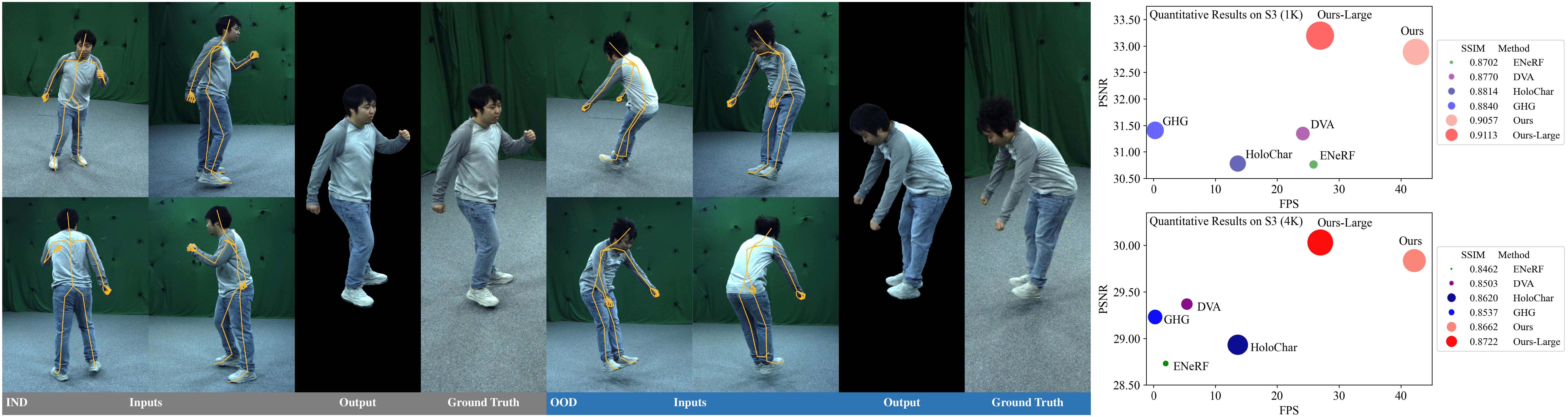}
        \vspace{-20pt}
        \captionof{figure}{
        We propose \textbf{Double Unprojected Textures} (\textbf{DUT}), a new method to synthesize photoreal 4K novel-view renderings in real-time.
        Our method consistently beats baseline approaches~\cite{lin2022efficient, remelli2022drivable,shetty2024holoported,kwon2024generalizable} in terms of rendering quality and inference speed.
        Moreover, it generalizes to, both, in-distribution (IND) motions, i.e. dancing, and out-of-distribution (OOD) motions, i.e. standing long jump.
        }
	\label{fig:futureTeas}
\end{center}
}]

\begin{abstract}
    Real-time free-view human rendering from sparse-view RGB inputs is a challenging task due to the sensor scarcity and the tight time budget.
To ensure efficiency, recent methods leverage 2D CNNs operating in texture space to learn rendering primitives.
However, they either jointly learn geometry and appearance, or completely ignore sparse image information for geometry estimation, significantly harming visual quality and robustness to unseen body poses.
To address these issues, we present \textbf{Double Unprojected Textures}, which at the core disentangles coarse geometric deformation estimation from appearance synthesis, enabling robust and photorealistic 4K rendering in real-time.
Specifically, we first introduce a novel image-conditioned template deformation network, which estimates the coarse deformation of the human template from a first unprojected texture.
This updated geometry is then used to apply a second and more accurate texture unprojection.
The resulting texture map has fewer artifacts and better alignment with input views, which benefits our learning of finer-level geometry and appearance represented by Gaussian splats.
We validate the effectiveness and efficiency of the proposed method in quantitative and qualitative experiments, which significantly surpasses other state-of-the-art methods.

\end{abstract}

\section{Introduction} \label{sec:introduction}
Sparse free-view human rendering concerns about rendering a virtual twin of a real human observed from (multiple) cameras into \textit{arbitrary} virtual view points.
Here, the ultimate pursuit is a fast, accurate, and robust algorithm, which captures human performance and reproduces realistic renderings in real time with minimal sensors.
Such an approach would have great potential in applications like immersive telepresence and it may completely change the way of how humans remotely communicate in the future.
While researchers have put immense efforts in solving this challenging task, the readily available solutions fall short in one or multiple ways, e.g., requiring dense sensors, limited photorealism, or slow runtime performance.
\par
Earlier works rely on dense camera setups to reconstruct high-quality geometry~\cite{collet2015high, guo2019relightables, neus2} and renderings~\cite{wu2020multi,mildenhall2020nerf}, which are typically not affordable to non-professional users.
Moreover, these methods generally fail under sparse camera assumptions, i.e. four or less cameras, due to inherent ambiguities and lack of observation.
To solve such ambiguities, some works~\cite{ARAH:ECCV:2022,gneuvox,sun2024metacap} learn priors from data, and perform expensive fine-tuning on novel images, making them inappropriate for live inference. 
Other works~\cite{shetty2024holoported, remelli2022drivable, Pang_2024_CVPR,kwon2024generalizable} efficiently predict rendering primitives in 2D texture space relying on a texture unprojection step~\cite{shetty2024holoported}.
However, they either jointly learn geometry and appearance from imperfect texture unprojections~\cite{remelli2022drivable,kwon2024generalizable} leading to reduced rendering quality and robustness or they completely ignore information readily available in the sparse views when predicting geometric deformations~\cite{shetty2024holoported} causing inconsistent wrinkles.
\par
In stark contrast, and to effectively utilize image clues, we \textit{explicitly} split human capture into coarse geometry reconstruction and fine-grained rendering primitives estimation.
We first leverage sparse image information to guide the geometry reconstruction, and then use this refined geometry to guide the learning of rendering primitives again conditioned on the sparse images. 
To ensure real-time capability, an integral part is to warp and fuse image information into a spatially aligned 2D space such that geometry as well as appearance can be effectively learned in 2D using lightweight 2D CNNs. 
\par
In detail, we propose \textbf{Double Unprojected Textures} (\textbf{DUT}), where a texture unprojection step is conducted twice for accurate geometry capture and realistic rendering.
It contains two stages: an efficient geometry estimation and a fine-grained Gaussian appearance prediction.
In stage one, we propose a novel image-conditioned template deformation approach. 
Significantly different from the motion-conditioned template deformation~\cite{habermann2021real} and pixel-aligned occupancy field~\cite{zheng2021deepmulticap}, our method utilizes local image features in texel space to efficiently predict human deformation on a canonicalized pose in 2D texture space, greatly improving robustness to out-of-distribution motions (see also Fig.~\ref{fig:futureTeas}).
More specifically, we firstly pose a human template mesh via linear blend skinning (LBS)~\cite{lewis2000pose} such that the geometry roughly aligns with the sparse images.
Then, our \textbf{first} texture unprojection converts image information, which contains coarse geometry information, into the canonical 2D texture space obtaining a coarse texture map.
Taking this map as input, we use a UNet~\cite{ronneberger2015u} to estimate the offsets of template vertices, which generates robust and smooth geometry.
In the second stage, we apply LBS to the \textit{deformed} template and perform the \textbf{second} texture unprojection. 
Notably, the deformed geometry better resembles the true surface and, thus, the texture map is more accurate with less artifacts and distortions.
This improved map can then be used as a conditioning input to our second UNet~\cite{ronneberger2015u} to estimate the Gaussian parameters.
Moreover, we found that geometry deformations may negatively affect the Gaussian appearance when not handled well.
Therefore, we introduce a Gaussian scale refinement where the template deformation guides the Gaussian scales.
\par
Our core contributions can be summarized as follows:
\begin{itemize} 
    \item A double texture unprojection strategy that decouples geometry recovery and photorealistic human rendering.
    \item An image-conditioned template deformation module that fully utilizes multi-view image clues in texel space.
    \item A Gaussian scale refinement that mitigates the negative impacts from geometric deformations to Gaussian scales.
\end{itemize} 
We compare our method against the prior state-of-the-art approaches on several benchmarks~\cite{habermann2021real,Pang_2024_CVPR,zheng2022structured} and show that DUT quantitatively and qualitatively outperforms prior works. 
Importantly, as our entire approach only relies on time-efficient 2D CNNs, we achieve an end-to-end real-time performance on a single GPU.

\section{Related Work} \label{sec:related}
\par 
\noindent\textbf{Template-based Performance Capture.}\label{subsec:rw_template}
Prior methods~\cite{gall2009motion,vlasic2008articulated,anguelov2005scape,bogo2015detailed, guo2021human} assume a static 3D scan or a parametric model of the person is provided, i.e. a 3D template, which can act as a strong prior for the typically ill-posed performance capture task.
This often results in improved performance and robustness compared to template-free methods.
In the context of human capture, human templates are firstly used for marker-less motion capture~\cite{theobalt2003parallel,liu2011markerless}.
Later, researchers investigated how to update the template geometry and motion simultaneously from multi-view cameras~\cite{de2008performance,Robertini:2016,sun2021neural}, monocular cameras~\cite{xu2018monoperfcap, habermann2019livecap, guo2021human}, or depth cameras~\cite{dou2016fusion4d,yu2017bodyfusion, yu2018doublefusion,xu2019unstructuredfusion}.
However, all these methods require expensive optimization and can easily fall into local minima.
In this regard, DeepCap~\cite{deepcap} was the first learning-based method for template-based human capture. 
It takes global image features to predict the skeletal motion and template deformation.
DDC~\cite{habermann2021real} further develops this idea to a motion-conditioned deformable avatar, widely used in animatable avatars~\cite{kwon2023deliffas,zhu2024trihuman,habermann2023hdhumans}.
Meanwhile, a series of neural implicit methods~\cite{tiwari2021neural, wang2021metaavatar,weng2022humannerf,guo2023vid2avatar} also deform canonical SDF fields with motion conditions.
Although such learning-based methods generate more stable results and achieve better accuracy, the global motion feature inputs limit their generalization ability to novel motions. 
Conversely, we propose a novel image-conditioned template deformation, which takes local image features in texel space as input and outputs more robust and accurate template deformations.
\par 
\noindent\textbf{Sparse-view Human Capture and Rendering.} \label{subsec:rw_sparse}
As a balance between high-quality dense-view camera rigs~\cite{collet2015high, guo2019relightables, shao2022diffustereo,zheng2024gpsgaussian}, i.e., 6 or more cameras, and lightweight monocular setups~\cite{saito2019pifu,Sengupta_2024_CVPR}, sparse-view camera settings, i.e. 4 or less cameras, serve as a good compromise for human capture and rendering.
Due to the highly under-constrained nature of sparse-view setups, early methods~\cite{theobalt2004combining,starck2007surface} exploit silhouette information and feature correspondence to reconstruct the visual hull~\cite{matusik2000image} or to deform a human template.
\citet{huang2018deep} combine CNNs with visual hull to predict occupancy fields from pixel-aligned features. 
DeepMultiCap~\cite{zheng2021deepmulticap} extends this idea with additional parametric model features.
Instead of occupancy values, a series of works estimates radiance fields~\cite{kwon2021neural}, volumetric primitives~\cite{remelli2022drivable}, texture maps~\cite{shetty2024holoported}, or 3D Gaussians~\cite{kwon2024generalizable} for improved rendering quality.
Others learn different priors~\cite{ARAH:ECCV:2022,gneuvox, sun2024metacap, huang2024tech} for superior fine-tuning performance, while it takes minutes to hours to run them.
In general, most of these works are far from real-time performance.
In stark contrast, our method is real-time capable while also achieving photorealistic rendering quality as our formulation can be entirely implemented using lightweight 2D CNNs.
\par 
\noindent\textbf{Real-time Human Capture and Rendering.} \label{subsec:rw_realtime}
Pioneering works~\cite{molet1996real, kalra1998real, horprasert1998real} formulated real-time human capture as a motion capture and animation task, but their animatable character renderings often lack realism.
With the aid of depth cameras, some approaches~\cite{orts2016holoportation,dou2017motion2fusion,tao2021function4d,jiang2022neuralhofusion} fuse RGBD information into sign distance fields~\cite{curless1996volumetric} to dynamically update geometry and appearance.
However, they are constrained by the shortcomings of depth cameras: high cost, low resolution, and lighting sensitivity.
At the same time, such methods are sensitive to fast body motion.
Monocular RGB methods~\cite{habermann2019livecap,li2020monocular,feng2022fof} estimate the template deformation or variants of pixel-aligned implicit functions~\cite{saito2019pifu} for human reconstruction.
They are consumer-friendly, but the result quality is far behind multi-view methods~\cite{remelli2022drivable,shao2022floren,shetty2024holoported}.
\par
The most closely related works are real-time sparse-view methods~\cite{remelli2022drivable,shetty2024holoported,kwon2024generalizable} that learn rendering primitives in texel space.
However, their method designs involve using a \textit{single} unprojection or multi-scaffolds~\cite{porumbescu2005shell}, which inherently limits their rendering quality, as geometry and appearance parameters are estimated jointly.
Instead, we decompose the human capture to coarse geometry estimation and fine-grained appearance modeling.
HoloChar~\cite{shetty2024holoported} also decomposes geometry and appearance. 
However, their geometry model is only conditioned on the motion signals (and not on image inputs), thereby resulting in weak generalization to novel images and novel motions.
Instead, we leverage image information at \textit{all} stages, i.e. geometry recovery and appearance synthesis, thus, greatly improving robustness and accuracy.
Moreover, compared to their single mesh representation, our Gaussian parameterization is capable of further learning fine-grained offsets.
\begin{figure*}[tb]
    \centering
	\includegraphics[width=0.98\linewidth]{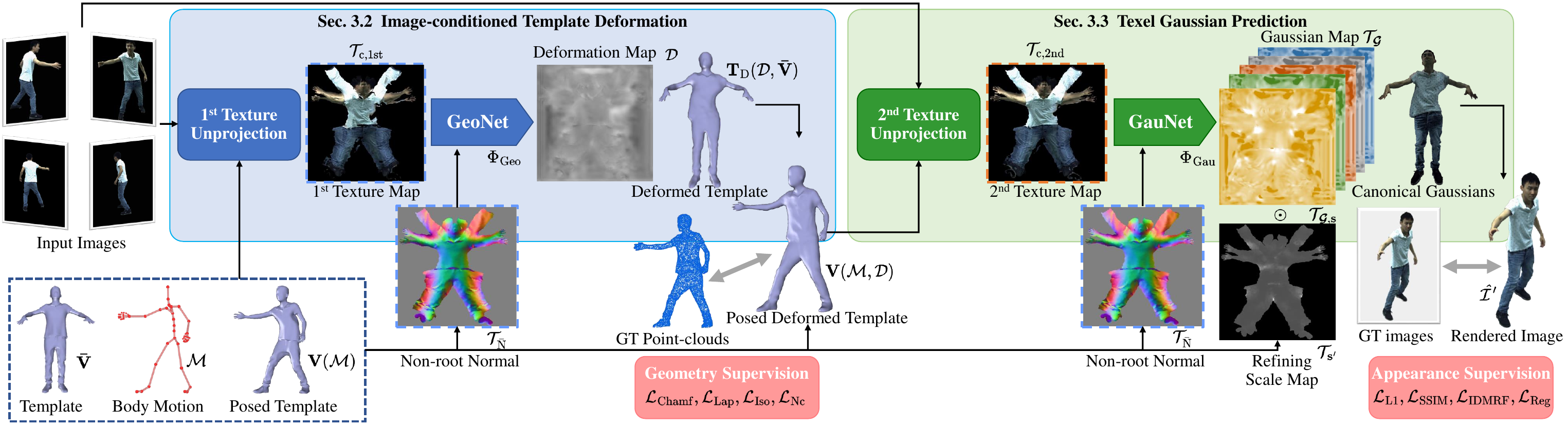}
    \vspace{-10pt}
	\caption
	{
        \textbf{Overview of DUT}.
        Given sparse-view images and respective motion, DUT predicts coarse template geometry and fine-grained 3D Gaussians.
        We first unproject images onto the posed template to obtain a texture map, which is fed into GeoNet to estimate deformations of the template in canonical pose.
        We then unproject images again onto the posed and deformed template to obtain a less-distorted texture map, which serves as input to our GauNet estimating 3D Gaussian parameters and undergoes scale refinement before splatting.
	}
    \vspace{-15pt}
	\label{fig:overview}
\end{figure*}

\section{Method} \label{sec:method}
Our goal is to capture the personalized human geometry and produce photo-realistic renderings from sparse-view RGB videos in real time. 
During training, we assume a collection of dense-view videos with ground truth foreground segmentations~\cite{lin2021real, kirillov2023segany} and skeletal motions~\cite{captury}. 
At inference, given segmented sparse-view ($1 \sim 4$) RGB videos of the subject and the corresponding skeletal motions, our method reconstructs the non-rigidly deforming geometry and renders photoreal free views, both, in real time (see also Fig.~\ref{fig:overview}).
\par 
In the following, we first recap the background knowledge in Sec.~\ref{subsec:preliminaries}.
In Sec.~\ref{subsec:template_deformation}, we propose a novel image-conditioned template deformation model, which first projects visual clues from input images into the texel space of a body template, and then learns the non-rigid surface deformation in texel space allowing to leverage efficient 2D CNNs.
With such deformed geometry, we perform a second unprojection to obtain a refined texture map, which better encodes the image information since projection artifacts caused by inaccurate geometry are significantly reduced.
Now, our GauNet (Sec.~\ref{subsec:gaussian_prediction}) takes this refined texture map to predict the 3D Gaussian parameters in texture space, which can be rendered into images by first transforming them to global space and then splatting them into screen space.
\subsection{Preliminaries}\label{subsec:preliminaries}
\noindent\textbf{Body Template.} A body template generally contains a base mesh paired with a skeleton. 
Given deformation parameters $\mathcal{D}$ and motion parameters $\mathcal{M}$ as inputs, we can non-rigidly deform the base mesh in the canonical pose and drive it to target pose through LBS~\cite{lewis2000pose}:
\begin{equation} \label{eq:template_transformtation}
\mathbf{V}(\mathcal{M},\mathcal{D}) = \mathbf{T}_{\mathrm{LBS}}(\mathcal{M})\mathbf{T}_{\mathrm{D}}(\mathcal{D}, \mathbf{\bar{V}}) 
\end{equation}
where $\mathbf{\bar{V}}\in \mathbb{R}^{ N_\mathrm{V} \times 3}$ denotes the vertices of the base mesh, the number of vertices is $N_\mathrm{V}$, $T_{\mathrm{LBS}}(\cdot)$ represents LBS transformation function that poses the vertices from the canonical to the world space, and $T_{\mathrm{D}}(\cdot)$ is used to deform the base mesh (detailed in Sec.~\ref{subsec:template_deformation}).
\par \noindent\textbf{3D Gaussians.} 
3DGS marries the idea of point splatting~\cite{zwicker2001surface} and point radiance~\cite{xu2022point} with a 3D Gaussian representation, which possesses the advantage of fast rasterization and realistic rendering. 
Each Gaussian is defined with a mean position  $\textbf{p}$ and covariance matrix $\boldsymbol{\Sigma}$
\begin{equation}
    G(p) = e^{-\frac{1}{2}(x-\textbf{p})^{T}\boldsymbol{\Sigma}^{-1}(x-\textbf{p})}
\end{equation}
where $x$ is the position in the Euclidean space, $\boldsymbol{\Sigma}=\mathbf{R}\mathbf{S}\mathbf{S}^{T}\mathbf{R}^{T}$, $\mathbf{S}\in\mathcal{R}_{+}^{3}$ denotes the scale matrix, and $\mathbf{R}\in SO(3)$ is the rotation matrix.
Each Gaussian can be parameterized as $\textbf{G}_{i} = \{ \textbf{p}_{i},\textbf{h}_{i},\textbf{s}_{i},\textbf{r}_{i},\boldsymbol{\alpha}_{i} \}$, including 3D positions in world space, spherical harmonics, scaling, rotations, and opacity values, respectively.
In the rendering phase, given the view matrix $\mathbf{W}$ and the Jacobian of the affine approximation of the projective transformation, $\mathbf{J}$, the 2D covariance matrix $\boldsymbol{\Sigma}^{'}$ is computed as $\boldsymbol{\Sigma}^{'}=\mathbf{J}\mathbf{W}\boldsymbol{\Sigma}\mathbf{W}^{T}\mathbf{J}^{T}$. 
The color of each pixel is computed by alpha blending $N$ sorted Gaussians, $\hat{C} = \sum_{n\in N}\textbf{c}_{n}\boldsymbol{\alpha}_{n}^{'}\prod_{m=1}^{n-1}(1-\boldsymbol{\alpha}_{m}^{'})$, where $\textbf{c}_{n}=\mathbf{H}(\textbf{h}_\mathrm{n},\mathbf{W},\textbf{p}_{\mathrm{n}})$ denotes the color of Gaussian $n$.
$\mathbf{H}(\cdot)$ converts spherical harmonics coefficients $\textbf{h}$ into RGB colors, and
$\boldsymbol{\alpha}_{n}^{'}$ is the opacity value of Gaussians in 2D.
\subsection{Image-conditioned Surface Deformation}\label{subsec:template_deformation}
In this stage, we decouple geometry recovery from appearance synthesis and focus on how to obtain better coarse geometry from sparse-view images while maintaining fast inference speed.
Previous template-based methods predict body deformations using global image features~\cite{deepcap} or motion-only features~\cite{habermann2021real, shetty2024holoported}. 
This is either not robust to out-of-distribution data or completely ignores relevant information provided in the sparse view images.
Motivated by texel-aligned feature methods~\cite{remelli2022drivable, kwon2024generalizable}, we fully exploit image clues to guide the template deformation in the texel space.
To bridge the gap between multi-view 2D images and the 3D human template, we perform a texture unprojection mapping image pixels into texel space. 
Then, the obtained texture map can serve as input to 2D CNNs~\cite{li2021survey} for efficient and effective deformation learning. 
\par \noindent\textbf{Texture Unprojection.} 
Unprojection is the inverse operation  of camera's perspective projection, 
which unprojects pixels from the 2D images back into the 3D space.
In the context of human performance capture, unprojection can be interpreted as mapping the 2D pixels to human geometry defined as a \textit{skinned and posed template}.
With the UV mapping~\cite{heckbert1986survey}, unprojected pixels are further mapped to the texture map $\mathcal{T}_\mathrm{c}$.
To fuse information across views while also taking geometry visibility into account, we compute a texel visibility map~\cite{shetty2024holoported,remelli2022drivable}
\begin{equation}
\mathcal{T}_\mathrm{v}^{i} = \mathcal{T}_\mathrm{v}^{\mathrm{angle},i} \wedge \mathcal{T}_\mathrm{v}^{\mathrm{depth},i} \wedge \mathcal{T}_\mathrm{v}^{\mathrm{mask},i}
\end{equation}
based on normal difference, depth difference, and segmentation masks for each camera view $i$ (see also supplemental document).
\par \noindent The final texture 
\begin{equation}
   \mathcal{T}_\mathrm{c} = 
    \begin{cases}
        \sum_{i} (\mathcal{T}_\mathrm{c}^{\mathrm{part}, i} \odot \mathcal{T}_{v}^{i} )/  \sum_{i} \mathcal{T}_{v}^{i}, & \text{if $\sum_{i} \mathcal{T}_{v}^{i} \neq 0 $} \\
        0, & \text{otherwise}
    \end{cases}
\end{equation}
is obtained by fusing partial textures $\mathcal{T}_\mathrm{c}^{\mathrm{part}, i}$ across views.
\par \noindent \textbf{Deformation Learning.}
By unprojecting textures onto the posed, but not deformed, body template, we obtain a heavily distorted texture map $\mathcal{T}_\mathrm{c, 1st}$, where image pixels are mapped to wrong texel positions on the texture map.
This is caused by geometric errors between the posed template and the real human surface.
However, $\mathcal{T}_\mathrm{c, 1st}$ still contains relevant information concerning the deformation state of the surface and we found it is sufficient to learn \textit{coarse} deformations.
Thus, we propose to train a geometry network, GeoNet $\Phi_{\mathrm{Geo}}$, which takes the texture map $\mathcal{T}_\mathrm{c, 1st}$ and normal map $\mathcal{T}_\mathrm{\bar{N}}$ of the posed template but without root rotation as inputs and outputs deformation parameters.
Unlike motion-conditioned methods~\cite{habermann2021real, shetty2024holoported}, our image features mitigate the one-to-many mapping issue~\cite{liu2021neural}, while the normal features provide additional information to ensure smoothness.
\par 
Previous approaches~\cite{habermann2021real, deepcap, tiwari2021neural} model cloth deformations with embedded graph deformations~\cite{sumner2007embedded} and vertex displacements.
In the pursuit of efficiency, we treat the deformations as vertex displacements upon the human template $\mathbf{\bar{V}}$ in canonical pose.
Our GeoNet $\Phi_{\mathrm{Geo}}$ learns the deformation maps $\mathcal{D}$ relative to body surface $\mathbf{\bar{V}}$:
\begin{align}
     \Phi_{\mathrm{Geo}}(\mathcal{T}_\mathrm{c, 1st}, \mathcal{T}_\mathrm{\bar{N}}) & = \mathcal{D} \\
    \mathbf{T}_{\mathrm{D}} (\mathcal{D},\mathbf{\bar{V}})&= (\pi_{uv}(\mathbf{\bar{V}}, \mathcal{D}) + \mathbf{\bar{V}}) \label{eq:deformed_template}
\end{align}
where $\pi_{uv}(\mathbf{\bar{V}},\mathcal{D})$ indexes the mesh vertices in the UV deformation map $\mathcal{D}$.
We then use Eq.~\ref{eq:template_transformtation} and motion $\mathcal{M}$ to pose the deformed geometry into world space geometry $\mathbf{V}(\mathcal{M},\mathcal{D})$.
\par \noindent \textbf{Supervision.}
We apply Chamfer distance against ground truth point-clouds (reconstructed by NeuS2~\cite{neus2}) to supervise the training of $\Phi_{\mathrm{Geo}}$. 
To reduce geometry artifacts, we also add a Laplacian loss~\cite{vollmer1999improved}, isometry loss, and normal consistency loss to regularize surface deformation
\begin{equation}
    \mathcal{L}_\mathrm{Geo} = \mathcal{L}_\mathrm{Chamf} + \lambda_\mathrm{Lap}\mathcal{L}_\mathrm{Lap} + \lambda_\mathrm{Iso}\mathcal{L}_\mathrm{Iso} + 
    \lambda_\mathrm{Nc}\mathcal{L}_\mathrm{Nc}
\end{equation}
where we set $\lambda_\mathrm{Lap}=1.0$, $\lambda_\mathrm{Iso}=0.1$, $\lambda_\mathrm{Nc}=0.001$ in our experiments. 
For templates with hands, we set $\lambda_\mathrm{Iso}=0.5$ to preserve the structure of hands.
\begin{figure}[tb]
	\includegraphics[width=0.98\linewidth] {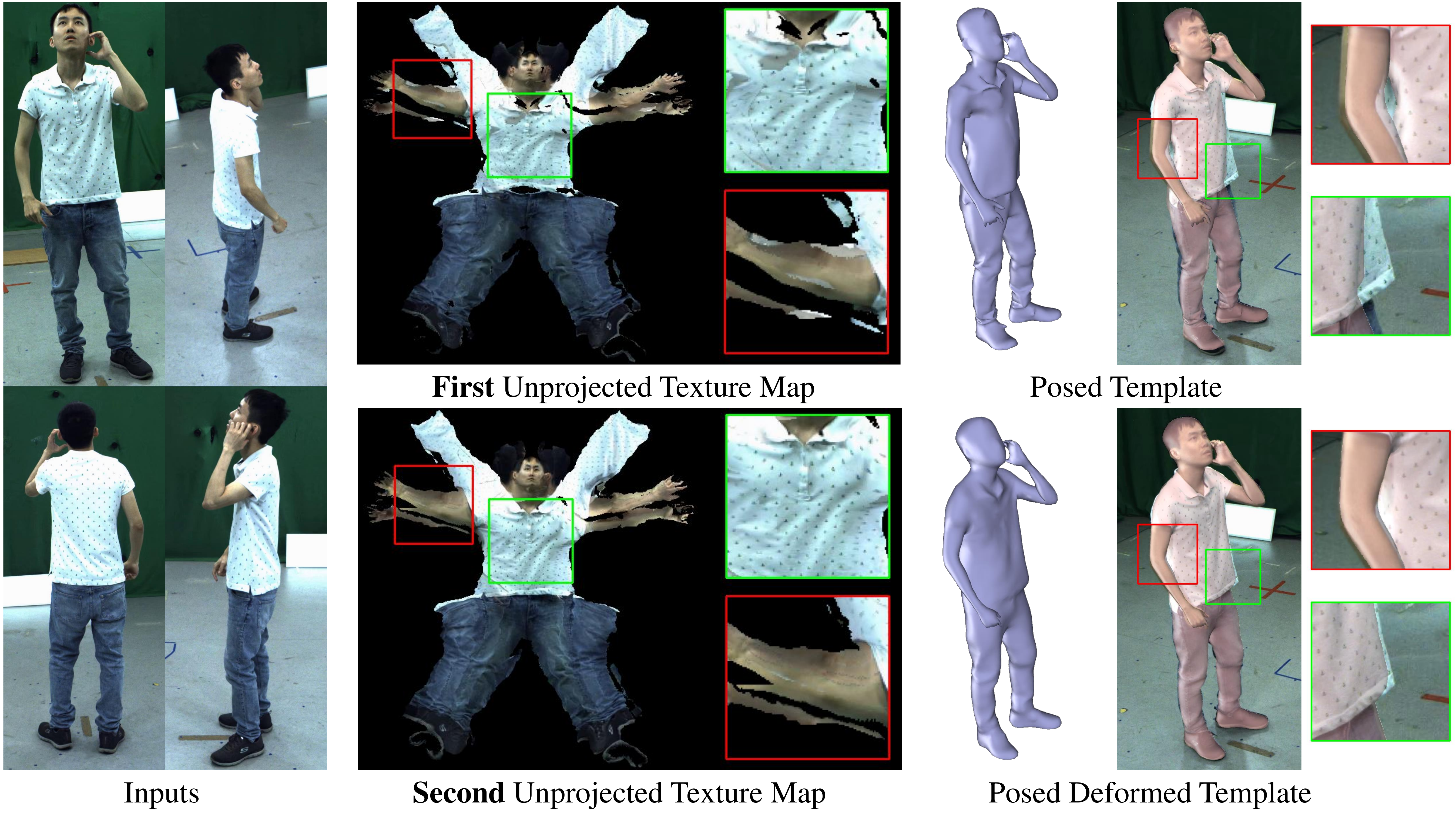}
   \vspace{-10pt}
	\caption
	{
        \textbf{Unprojected Texture Maps}.
        Performing a second texture unprojection using the deformed template leads to less ghosting artifacts and better geometric alignment.
	   }
       \vspace{-15pt}
	\label{fig:unprojectionmap}
\end{figure}

\subsection{Gaussian Appearance Prediction}\label{subsec:gaussian_prediction}
With GeoNet, we can capture a coarse geometry of the human, which approximately aligns with the body surface but it lacks detailed geometry and appearance.
To recover the fine-grained human details, we adopt 3D Gaussians ~\cite{kerbl20233d} as rendering primitives, due to their ability to model realistic colors, geometry displacements, and the support for fast rasterization. 
Meanwhile, having coarse geometry from the previous stage significantly improves the rendering quality of the learned Gaussians as we outline in the following.
\par 
Inspired by previous methods~\cite{Pang_2024_CVPR, li2024animatable, kwon2024generalizable} that use normal maps, position maps, or partial texture maps to estimate 3D Gaussians, we propose GauNet $\Phi_{\mathrm{Gau}}$, which converts texture and normal maps into 3D Gaussian parameters $\{\boldsymbol{\mathcal{G}}_{i}, {i}\in[1,N_\mathrm{G}]\}$ in texel space, where each texel corresponds to a Gaussian.
Different from the position $\textbf{p}_{i}$ of the original Gaussian $\textbf{G}_{i}$, we estimate 3D displacements $\textbf{d}_{i}$ to the canonical \textit{deformed} template defined in Eq.~\ref{eq:deformed_template}.
In total, the network estimates the parameters $\boldsymbol{\mathcal{G}}_{i} = \{ \textbf{d}_{i},\textbf{h}_{i},\textbf{s}_{i},\textbf{r}_{i},\boldsymbol{\alpha}_{i} \}$ for each texel.
As discussed, if the coarse geometry is inaccurate, the unprojected texture will be severely distorted. 
Directly estimating Gaussians from such geometry and unprojected texture is not effective, as the coarse geometry and Gaussian parameters are coupled.
In stark contrast to previous methods~\cite{remelli2022drivable, shetty2024holoported, kwon2024generalizable}, we conduct texture unprojection \textit{again} to obtain a less distorted texture map $\mathcal{T}_\mathrm{c, 2nd}$, where image pixels are unprojected  onto the \textit{posed} and \textit{deformed} template from stage 1.
In this way, we relieve the burden of GauNet from learning large vertex displacements to learning small displacements. 
Moreover, the predicted color $\textbf{c}_\mathrm{i}$ has a stronger relation to the color inputs (see also Fig.~\ref{fig:unprojectionmap}).
The prediction of Gaussian maps $\mathcal{T}_{\boldsymbol{\mathcal{G}}}$ can be formulated as:
\begin{align}
     \mathcal{T}_{\boldsymbol{\mathcal{G}}} &= \Phi_{\mathrm{Gau}}(\mathcal{T}_\mathrm{c, 2nd}, \mathcal{T}_\mathrm{\bar{N}})\\
     \textbf{G} &= \mathcal{T}_{\boldsymbol{\mathcal{G}}}[\mathcal{M}_{\boldsymbol{\mathcal{G}}}]
\end{align}
$\mathcal{M}_\mathcal{G}$ is a fixed mask map of valid Gaussians in texel space. 
$[.]$ is the index operation. 
We will omit $\mathcal{M}_\mathcal{G}$ for readability.
\par
To pose the Gaussians from canonical pose space to the world space, we render the LBS transformations map $\mathcal{T}_\mathrm{LBS}$ and the deformed base geometry map $\mathcal{T}_{\mathbf{T}_{\mathrm{D}}(\mathcal{D}, \mathbf{\bar{V}})}$ for each motion $\mathcal{M}$. 
Given camera's view matrix $\mathbf{W}$ and the Gaussian renderer $R$, we render a novel view $\mathcal{\hat{I}}$ as:
\begin{align}
     \textbf{p} &= \mathcal{T}_\mathrm{LBS}(\mathcal{T}_{{\boldsymbol{\mathcal{G}}},\textbf{d}} + \mathcal{T}_{\textbf{T}_{\mathrm{D}}(\mathcal{D}, \mathbf{\bar{V}})}) \\
     \textbf{r} &=  \mathcal{T}_\mathrm{LBS}\mathcal{T}_{{\boldsymbol{\mathcal{G}}},\textbf{r}} \\
     \mathcal{\hat{I}} & = R(\mathbf{W}, \textbf{G}(\textbf{p},\textbf{h},\textbf{s},\textbf{r},\boldsymbol{\alpha}))
\end{align}
where $\mathcal{T}_{\mathcal{G},\mathrm{\chi}}$ denotes the features $\mathrm{\chi}$ of $\mathcal{T}_{\mathcal{G}}$, $\mathrm{\chi}$ is one of features $\{\textbf{d},\textbf{h},\textbf{s},\textbf{r},\boldsymbol{\alpha} \}$, i.e. $\textbf{h}= \mathcal{T}_{{\boldsymbol{\mathcal{G}}},\textbf{h}}, \textbf{s}= \mathcal{T}_{{\boldsymbol{\mathcal{G}}},\textbf{s}}$, and $\boldsymbol{\alpha}=\mathcal{T}_{{\boldsymbol{\mathcal{G}}},\boldsymbol{\alpha}}$.

\par \noindent\textbf{Gaussian Scale Refinement.} 
Though the above prediction and rendering process already achieves impressive rendering results, it could produce noticeable scale-related artifacts at certain body poses.
We found this is due to the fact that, for some body motions, LBS can induce significantly scale changes, i.e. triangles are significantly stretched, on the human body, which makes it hard for $\Phi_{\mathrm{Gau}}$ to learn appropriate scales.
Thus, we introduce a Gaussian scale refinement to refine the scale parameters of the Gaussians explicitly.
Specifically, we compute the refining scales based on the scaling ratios from posed template to canonical template for all nearby edges, and select the maximum ratios as the refining scales. 
We also clamp refining scales to ensure them being equal to or greater than 1.0, since decreasing scales has the risk of vanishing gradients.
The refining scales can be computed as
\begin{equation}
    \textbf{s}'_{i} = \mathbf{max}( \mathbf{max}( \{ \textbf{s}'_{{i}, {j} } , {j} \in \mathcal{E}_{i} \})   , 1.0), {i}\in [1,N_\mathrm{V}],
\end{equation}
where $\mathcal{E}_{i}$ is the edge set of vertex ${i}$, $\mathrm{s}'_{{i}, {j}}$ is the scaling ratio of the edge ${j}$ of vertex ${i}$. 
We also render refining scales $\mathrm{s}'$ in to refining scale map $\mathcal{T}_{\mathrm{s}'}$. 
The refined scales and rendering can be represented as:
\vspace{-5pt}
\begin{align}
      \textbf{s}'  &=  \mathcal{T}_{\textbf{s}'} \odot \mathcal{T}_{{\boldsymbol{\mathcal{G}}},\textbf{s}} \\
     \mathcal{\hat{I}}' & = R(\mathbf{W}, \textbf{G}(\textbf{p},\textbf{c},\textbf{s}',\textbf{r},\boldsymbol{\alpha}))
\end{align}
\par \noindent \textbf{Supervision.}
The supervision of GauNet involves an L1 loss, an SSIM loss~\cite{wang2004image}, and an IDMRF loss~\cite{wang2018image} for image reconstruction as well as a geometric regularization loss:
\begin{equation}
    \mathcal{L}_\mathrm{Gau} = \mathcal{L}_\mathrm{L1} + \lambda_\mathrm{SSIM}\mathcal{L}_\mathrm{SSIM} + \lambda_\mathrm{IDMRF}\mathcal{L}_\mathrm{IDMRF} + \lambda_\mathrm{Reg}\mathcal{L}_\mathrm{Reg}
\end{equation}
where we set $\lambda_\mathrm{SSIM}=0.1$, $\lambda_\mathrm{IDMRF}=0.01$ and $\lambda_\mathrm{Reg}=0.005$ in all our experiments with $\mathcal{L}_\mathrm{Reg} = \|\mathcal{T}_{\boldsymbol{\mathcal{G}},\textbf{d}} \|_{2}^{2}$. For large Gaussian texture map, we set $\lambda_\mathrm{Reg}=0.01$.
\begin{figure*}[t]
        \centering
	\includegraphics[width=0.95\linewidth]{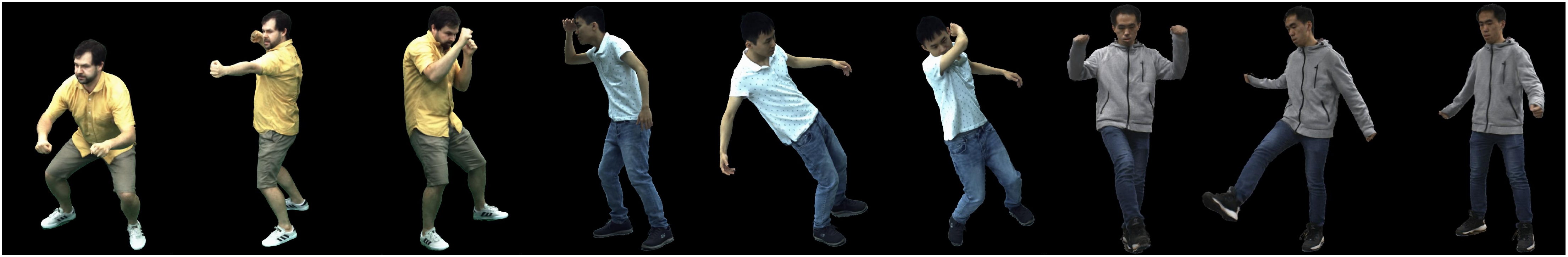}
     \vspace{-10pt}
	\caption
	{
        \textbf{Qualitative Results}.
        Here, we demonstrate novel-view rendering results on novel poses at 4K or full resolutions. 
        Note that DUT faithfully captures cloth wrinkles, body details on hands and faces.
        Besides, it is robust to diverse poses and self occlusions.
	}
    \vspace{-12pt}
	\label{fig:qualitative_results}
\end{figure*}

\begin{figure*}[tb]
        \centering
	\includegraphics[width=0.95\linewidth]{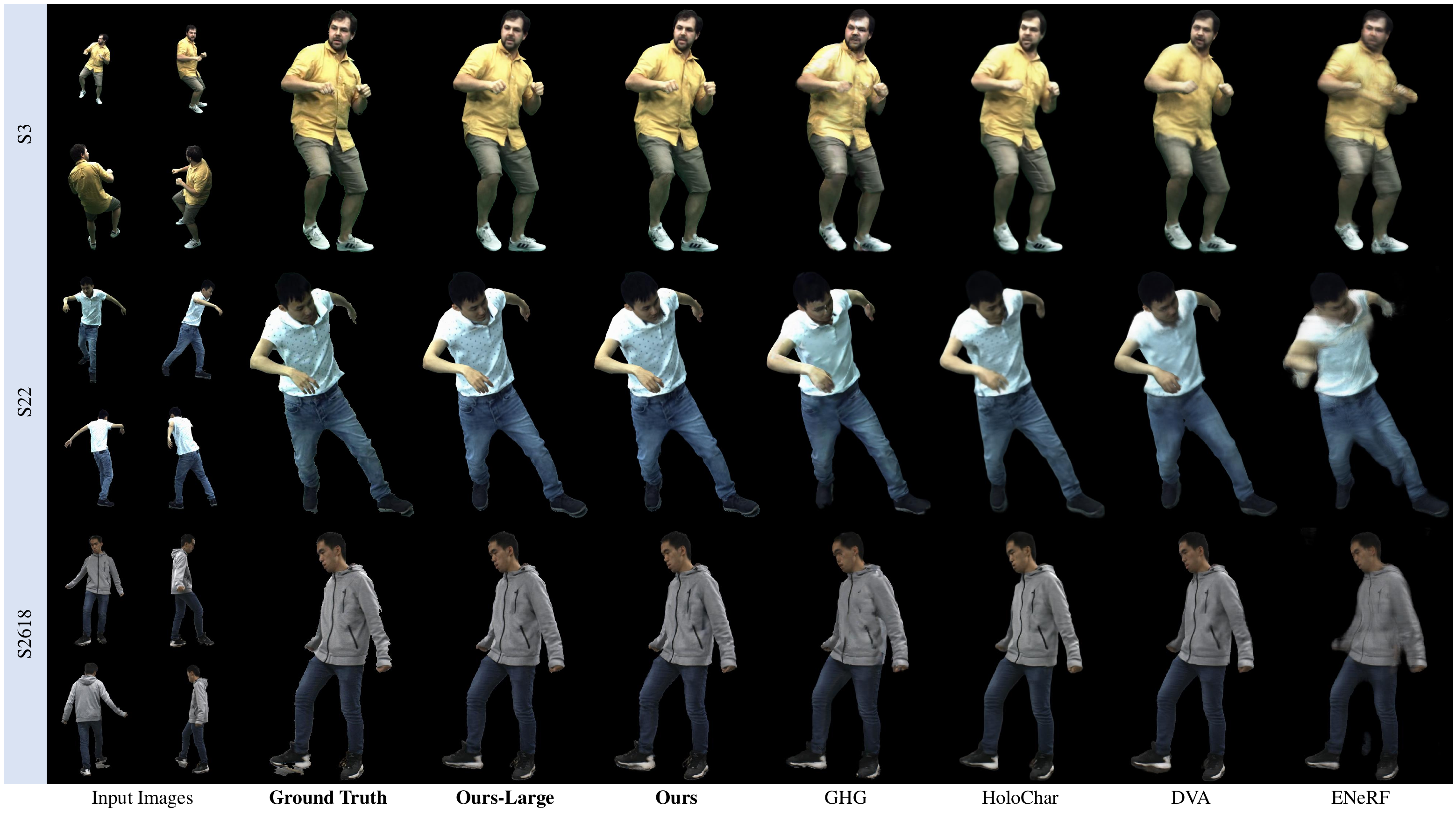}
        \vspace{-10pt}
	\caption
	{
        \textbf{Qualitative Comparison.}
        Given sparse-view images with novel poses, our method captures sharper and more faithful appearance details including facial expressions, hand gestures, and cloth wrinkles at 4K or full resolution, compared to prior works, i.e. GHG~\cite{kwon2024generalizable}, HoloChar~\cite{shetty2024holoported}, DVA~\cite{remelli2022drivable}, and ENeRF~\cite{lin2022efficient}.
        Notably, our method is also faster than previous methods.
	}
    \vspace{-20pt}
	\label{fig:comparison}
\end{figure*}

\section{Results} \label{sec:results}
\noindent\textbf{Implementation Details.}
In stage I, we set the texture size to 256 and train $\Phi_{\mathrm{Geo}}$ for 150k steps using AdamW~\cite{loshchilov2017decoupled} with a batch size of 1 and a learning rate of 2e-4.
For $\Phi_{\mathrm{Gau}}$ in stage II, we follow ASH~\cite{Pang_2024_CVPR} having 15k warmup steps and train till 680k steps on 1K resolution images.
Afterwards, we finetune the network on 4K images till 1.97M total steps.
We use AdamW~\cite{loshchilov2017decoupled} with a batch size of 1 and a learning rate of 1e-4 and decrease the learning rate to 1e-5 during step 650k to 680k and 1.9M to 1.97M.
The texture size is 256 for `Ours' and 512 for `Ours-Large'.
For runtime measurements, we use a PC with a single RTX3090 GPU and an Intel i9-10900X CPU.
Please refer to the supplementary material for more details.
\par \noindent\textbf{Datasets.}
We perform experiments on subjects from DynaCap~\cite{habermann2021real}, ASH~\cite{Pang_2024_CVPR}, and THUman4.0~\cite{zheng2022structured}, which offer dense-view video sequences, human templates, and corresponding body motions.
For S2618 from THUman4.0, we use the deformed SMPL model~\cite{SMPL:2015} as the human template.
To examine the performance on out-of-distribution motions, we collect a subject with dense-view cameras.
\par \noindent\textbf{Qualitative Results.} 
Fig.~\ref{fig:qualitative_results} demonstrates the qualitative results of our method on testing sequences.
Our method achieves high-fidelity novel-view renderings of humans at 4K resolution, while it faithfully recovers human geometry and appearance details for diverse subjects and motions.
\begin{table}[t]
    \footnotesize
   \centering
   \setlength\tabcolsep{1pt}
    \scalebox{0.98}{
        \begin{tabular}{c|c|c|c|c|c|c}
        \hline 
        \multirow{2}{*}{\textbf{Method}} & \textbf{PSNR} $\uparrow$ & \textbf{SSIM} $\uparrow$ & \textbf{LPIPS} $\downarrow$ & \textbf{PSNR} 
        $\uparrow$ & \textbf{SSIM} $\uparrow$ & \textbf{LPIPS} $\downarrow$   \\ 
        \cline{2-7}
        & \multicolumn{3}{c|}{\textbf{S3 - 1K}} & \multicolumn{3}{c}{\textbf{S3 - 4K}}\\
       \hline ENeRF ~\cite{lin2022efficient} & 30.7611 & 0.8702 & 0.0943 & 28.7297 & 0.8462 & 0.1590 \\
        \hline DVA ~\cite{remelli2022drivable} & 31.3487 & 0.8770 & 0.0910 & \bronze{29.3681} & 0.8503 & 0.1655 \\
        \hline 
        HoloChar ~\cite{shetty2024holoported} & 30.7795 & 0.8814 & \bronze{0.0841} & 28.9313 & \bronze{0.8620} & \bronze{0.1549} \\
        \hline 
        GHG ~\cite{kwon2024generalizable}& \bronze{31.4086} & \bronze{0.8840} & 0.0907 & 29.2304 & 0.8537 & 0.1551 \\
        \hline 
        \textbf{Ours} & \silve{32.8897} & \silve{0.9057} & \silve{0.0612} & \silve{29.8369} & \silve{0.8662} & \silve{0.1389} \\
        \hline 
        \textbf{Ours-Large} & \gold{33.1960} & \gold{0.9113} & \gold{0.0560} & \gold{30.0311} & \gold{0.8722} & \gold{0.1322}\\
        \hline 
         & \multicolumn{3}{c|}{\textbf{S22 - 1K}} & \multicolumn{3}{c}{\textbf{S22 - 4K}}\\
         \hline ENeRF ~\cite{lin2022efficient} & 32.7524 & 0.8721 & 0.0853 & 30.2329 & 0.8325 & 0.1703\\
        \hline DVA ~\cite{remelli2022drivable} & \bronze{33.7842} & \bronze{0.8887} & \bronze{0.0664} & \gold{31.2019} & 0.8448 & 0.1711 \\
        \hline 
        HoloChar ~\cite{shetty2024holoported} & 32.2055 & 0.8850 & 0.0775 & 30.1391 & \silve{0.8533} & 0.1691\\
        \hline 
        GHG ~\cite{kwon2024generalizable}& 33.3187 & 0.8847 & 0.0846 & \silve{30.8346} & 0.8448 & \bronze{0.1675}\\
        \hline 
        \textbf{Ours} & \silve{34.1119} & \silve{0.9066} & \silve{0.0568} & 30.6427 & \bronze{0.8480} & \silve{0.1387} \\
        \hline 
        \textbf{Ours-Large} & \gold{34.2425} & \gold{0.9113} & \gold{0.0531} & \bronze{30.8126} & \gold{0.8565} & \gold{0.1290}\\
        \hline

        \hline 
         & \multicolumn{3}{c|}{\textbf{S2618 - Half Res}} & \multicolumn{3}{c}{\textbf{S2618 - Full Res}}\\
         \hline ENeRF ~\cite{lin2022efficient} & 29.6588 & 0.8761 & 0.0989 & 27.6079 & 0.8439 & 0.1537 \\
        \hline DVA ~\cite{remelli2022drivable} & \bronze{30.8135} & \bronze{0.8917} & \bronze{0.0774} & \bronze{29.2767} & 0.8611 & \silve{0.1295} \\
        \hline 
        HoloChar ~\cite{shetty2024holoported} & 28.3811 & 0.8722 & 0.1132 & 28.1441 & \bronze{0.8666} & 0.1602\\
        \hline 
        GHG ~\cite{kwon2024generalizable}& 29.6830 & 0.8831 & 0.1059 & 28.4991 & 0.8597 & 0.1587\\
        \hline 
        \textbf{Ours} & \silve{31.2413} & \silve{0.9101} & \silve{0.0747} & \silve{29.3799} & \silve{0.8719} & \bronze{0.1355} \\
        \hline 
        \textbf{Ours-Large} & \gold{31.5920} & \gold{0.9179} & \gold{0.0678} & \gold{29.6887} & \gold{0.8815} & \gold{0.1235}\\
        \hline
        \end{tabular}
        }
          \vspace{-5pt}
\caption{
\textbf{Quantitative Comparison.}
We evaluate our approach on 3 subjects, i.e. S3~\cite{habermann2021real}, S22~\cite{Pang_2024_CVPR}, and S2618~\cite{zheng2022structured} in terms of rendering quality and consistently outperform prior works for both variants of our method, i.e., \textit{Ours} and \textit{Ours-Large}.}
\vspace{-15pt}
\label{tab:quantitative}	
\normalsize
\end{table}

\begin{figure}[tb]
	\includegraphics[width=0.98\linewidth]{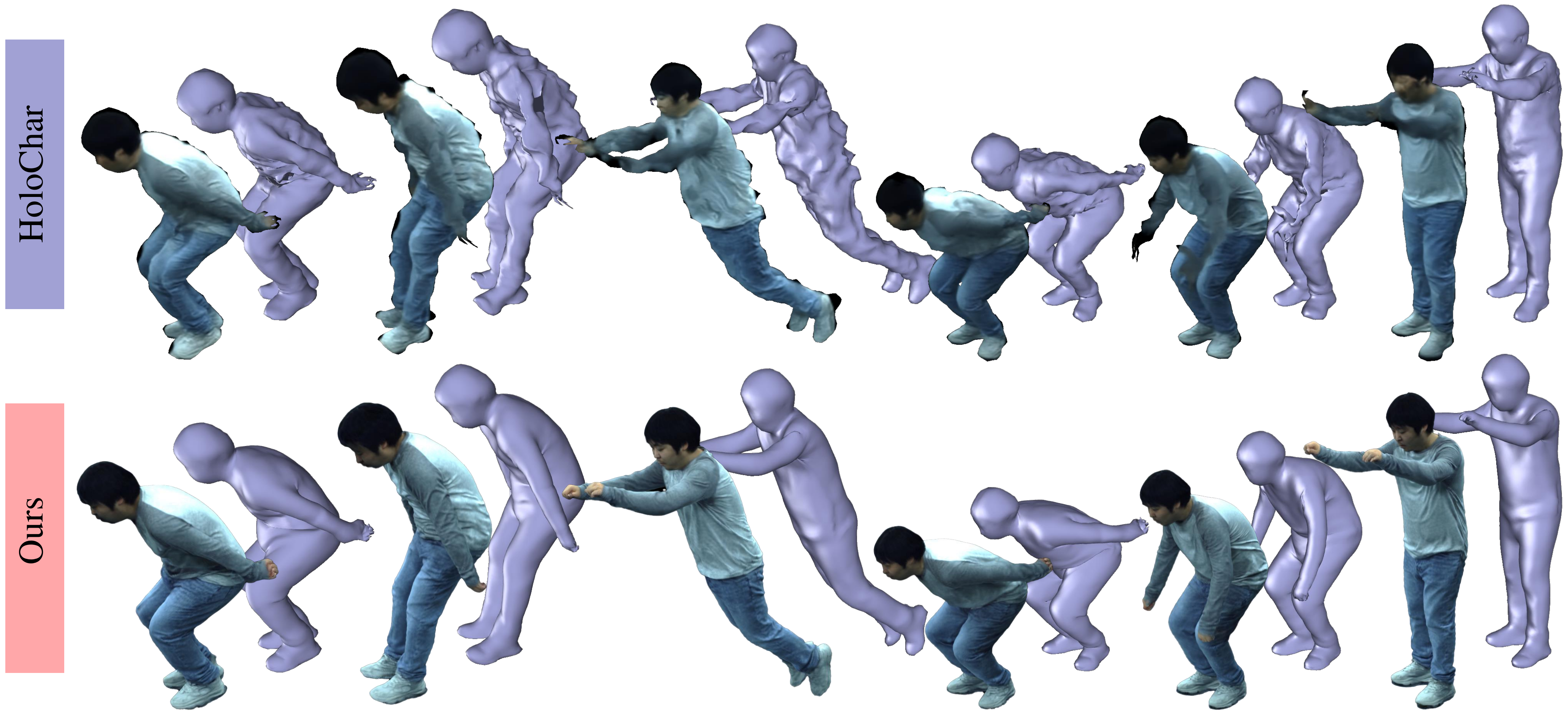}
     \vspace{-10pt}
	\caption
	{
        \textbf{Qualitative Comparison}.
         Compared to HoloChar~\cite{shetty2024holoported}, our method successfully generalizes to out-of-distribution (OOD) motions, e.g. long standing jump.
	}
        \vspace{-20pt}
	\label{fig:ood}
\end{figure}

\par \noindent\textbf{Evaluation Protocol.}
For comparisons, we train all baseline methods and variants of our method on the respective training frames and views.
For S3 and S22, we sample every 10\nth frame using both 1K and 4K resolution.
For S2618 from THUman4.0~\cite{zheng2022structured}, we include condition views for supervision and set a frame sampling rate of 3, due to its limited camera number and frame number.
In the ablations, we train all variants at 4K resolution.
We compute metrics on every 10\nth frame of the testing sequences on four isolated testing views.
In terms of geometry, we report Chamfer distance (CD), surface regularity (SR)~\cite{guillard2024latent}, and self-intersection (SI)~\cite{jung2004self} to evaluate the accuracy and smoothness of the recovered geometry.
For renderings, we report PSNR, SSIM~\cite{wang2004image}, and LPIPS~\cite{zhang2018unreasonable} to evaluate photometric consistency, structure similarity, and perceptual quality.
\subsection{Comparison} \label{sec:comparison}
\noindent\textbf{Competing Methods.} 
We compare our method to state-of-the-art sparse-view human rendering approaches that have fast or real-time inference speed.
DVA~\cite{remelli2022drivable} and HoloChar~\cite{shetty2024holoported} are designed for real-time holoporation, where they estimate volumetric primitives~\cite{remelli2022drivable} or neural texture maps~\cite{shetty2024holoported} from unprojected texture maps. 
Differently, DVA learns geometry and rendering parameters jointly, while HoloChar firstly estimate a coarse template deformation from motion signals only and then learns textures without updating the geometry.
GHG~\cite{kwon2024generalizable} is a fast and generalizable human rendering method, which predicts 3D Gaussians from unprojected textures on multi-scaffolds.
ENeRF~\cite{lin2022efficient} is a real-time generalizable neural rendering method for general scenes.
It uses cascade cost volumes to accelerate point sampling for volumetric renderings. 
\par \noindent\textbf{Comparison.}
We show detailed qualitative and quantitative comparisons to competing methods in Fig.~\ref{fig:comparison} and Tab.~\ref{tab:quantitative}.
ENeRF~\cite{lin2022efficient} relies on small camera baselines for effective feature matching, which is not feasible under the challenging sparse-view setup. 
Thus, it generates severe artifacts and performs significantly worse than our proposed approach.
GHG~\cite{kwon2024generalizable} neglects the coarse template geometry estimation via predicting 3D Gaussians on multiple shell maps around the base template. 
Such multi-level representation has noticeable artifacts on body and face.
DVA~\cite{remelli2022drivable} jointly learns the deformation and appearance parameters of volumetric primitives. 
Their coupling of geometry and appearance learning makes it hard to capture shaper details. %
HoloChar~\cite{shetty2024holoported} has a motion-conditioned deformable template as coarse geometry. 
However, they do not refine the template when learning neural textures, leading to misalignment between predicted wrinkles and real images. 
Besides, the neural textures are less effective for rendering 4K images.
In contrast, our method faithfully captures hands, face, and cloth wrinkles achieving consistently better visually results.
Regarding SSIM and LPIPS, there is a clear improvement between our method and baseline methods, which confirms that our method produces more accurate and faithful renderings.
\begin{figure}[tb]
	\includegraphics[width=0.99\linewidth]{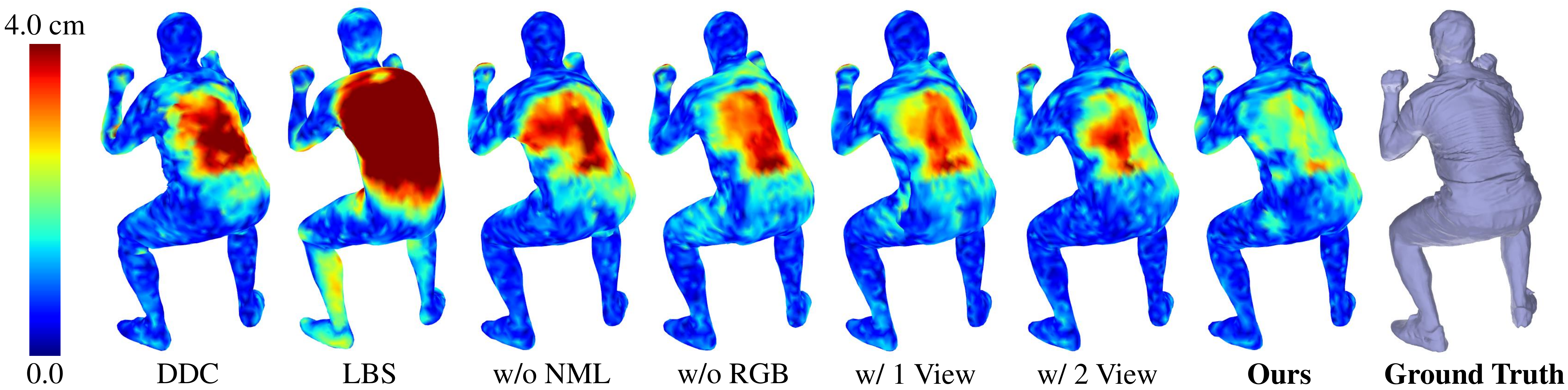}
     \vspace{-5pt}
	\caption
	{
        \textbf{Qualitative Ablation}.
        Here, we compare the performance of our design choices and of baseline methods, DDC~\cite{habermann2021real} and LBS~\cite{lewis2000pose}, in terms of template deformations.
        Our full model, using, both, RGB and normal features, achieves better accuracy.
	}
        \vspace{-10pt}
	\label{fig:ablation_geo}
\end{figure}

\begin{table}[t]
  \centering
      \small
          \scalebox{0.9}{
    \begin{tabular}{|c|c|c|c|c|c|}
    \hline
    \textbf{Methods} & \textbf{Backbone} & \textbf{Tex Res} & \textbf{CD}$\downarrow$ & \textbf{SR}$\downarrow$ & \textbf{SI} $\downarrow$ \\
    \hline 
     DDC & GCN & - & 1.173	& 0.388 &	1.162  \\
    \hline
     LBS & - & - & 2.117  	& \gold{0.236} &	\gold{0.768}  \\
    \hline
     w/o RGB & CNN & 256 & 1.150 	& \bronze{0.360} &	0.895  \\
     w/o NML  & CNN & 256 & 1.117 &	0.387 & 0.869 \\
     w/ 1 View  & CNN & 256 & \bronze{1.109} &	\silve{0.356 }	& \bronze{0.822} \\
     w/ 2 View  & CNN & 256 & \silve{1.096} &	0.365 &	0.881 \\
    \textbf{Ours} & CNN & 256 & \gold{1.070} & 0.374 & \silve{0.810} \\
    \hline
    \end{tabular}
    }
  \vspace{-5pt}
  \caption{
  \textbf{Quantitative Ablation.}
  Here, we study the influence of image features, normal features, number of views, and network architectures on the template deformation.
  Our full model with, both, sparse-view image clues and normal features achieves better accuracy while maintaining a smooth geometry.
  }
  \vspace{-5pt}
  \label{tab:ablation_geo}
  \normalsize
\end{table}

\begin{figure*}[tb]
        \centering
	\includegraphics[width=0.95\linewidth]{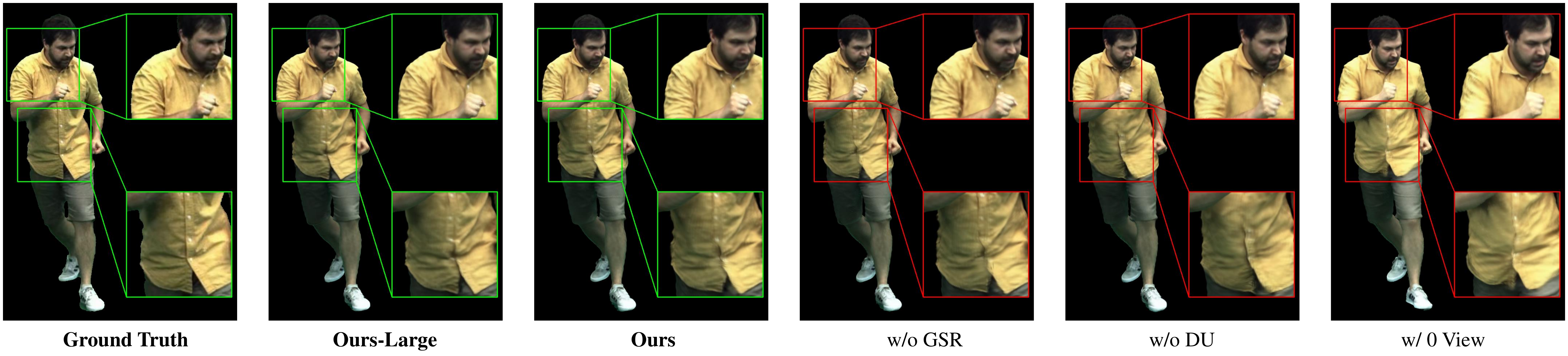}
        \vspace{-10pt}
	\caption
	{
        \textbf{Qualitative Ablation.}
        We evaluate the design choices of texel Gaussian prediction module. 
        Compared to baseline variants, our full method generates better visual results.
        Ours-Large achieves even better performance, at the cost of slower inference speed.
	}
    \vspace{-15pt}
	\label{fig:ablation_texture}
\end{figure*}

\begin{table}[t]
  \centering
      \small
          \scalebox{0.9}{
    \begin{tabular}{|c|c|c|c|c|c|}
    \hline
    \textbf{Methods} & \textbf{Tex Res} &  \textbf{PSNR} $\uparrow$ & \textbf{SSIM} $\uparrow$ & \textbf{LPIPS} $\downarrow$  \\

    \hline
     w/ 0 View &   256 & 25.1546	& 0.7874 &	0.1848  \\
     w/ 1 View &   256 & 27.9061	& 0.8340 &	0.1579  \\
     w/ 2 View &   256 & 29.3016	& 0.8568 &	0.1436   \\
    \hline
     w/ `w/o RGB Geo' &   256 & 29.7097	& 0.8638 &	0.1402  \\
     w/ `w/o NML Geo' &   256 & 29.6742	& 0.8640  &	0.1397  \\
     \hline
     w/o DU &   256 & 29.4185	& 0.8543 &	0.1452  \\
     w/o GSR &   256 & \bronze{29.8102}	& \bronze{0.8657} &	\bronze{0.1390}  \\
     Ours &  256 & \silve{29.8369} & \silve{0.8662} & \silve{0.1389} \\
    \hline
     Ours-Large &  512 & \gold{30.0311}	& \gold{0.8722} &	\gold{0.1322}  \\
    \hline
    \end{tabular}
    }
  \vspace{-5pt}
  \caption{
  \textbf{Quantitative Ablation.}
  We evaluate how the proposed components, number of views, template differences and texture resolutions affect the appearance module's performance at 4K resolution.
  The ``Double Unprojection'' (DU) and ``Gaussian Scale Refinement'' (GSR) contribute to the final results in our full model.
  }
  \vspace{-10pt}
  \label{tab:ablation_texture}
  \normalsize
\end{table}

\par \noindent\textbf{Comparison on Out-of-Distribution Motions.}
We compare our method with HoloChar~\cite{shetty2024holoported} on completely out-of-distribution (OOD) motions to evaluate their robustness. 
Thanks to the design of our image-condition template module, our method robustly captures geometry and appearance on OOD motions (Fig.~\ref{fig:ood}).
Since HoloChar solely relies on motion conditions, it is less robust to OOD motions.
This is also quantitatively confirmed as our method achieves a PSNR score of $\textbf{32.64}$ compared to $30.22$ for HoloChar.
\subsection{Ablation Studies} \label{sec:ablation}
\noindent\textbf{Image-conditioned Template Deformation.} 
In Fig.~\ref{fig:ablation_geo} and Tab.~\ref{tab:ablation_geo}, we perform an ablation study on the geometry module and investigate the influence of input features, view numbers, and network architectures.
The naive baseline `LBS' removes the deformation on the template.
While skinning-based deformation is smoother, it cannot recover the true geometry, confirming the need for our deformation network.
DDC~\cite{habermann2021real} and `w/o RGB', both, only take normal features as inputs.
Again, we find our design, i.e. using image conditioning, to perform superior compared to these baselines.
Only providing RGB features instead, referred to as `w/o NML', outputs higher accuracy than `w/o RGB' but the surface is less smooth.
Our full method achieves best capture accuracy while also maintaining a smooth and regular surface.
Besides, we can witness a gradual performance drop when reducing the number of input views.
\par \noindent\textbf{Texel Gaussian Prediction.}
As shown in Fig.~\ref{fig:ablation_texture}, our method with Gaussian scale refinement produces less stripes-like artifacts compared to `w/o GSR'.
Without double unprojection, the appearance module will struggle to capture high-frequency details, i.e., wrinkles on the arm.
If we remove RGB inputs in, both, our geometry module and our appearance module, referred to as `w/ 0 View', our method effectively turns into an animatable avatar method.
However, it loses the ability to faithfully reproduce the true appearance and geometry as this is a common limitation for animatible approaches.
If we increase the resolution of the Gaussian texture map from 256 to 512, our method obtains even higher fidelity renderings at the cost of reduced inference speed.
The quantitative results in Tab.~\ref{tab:ablation_texture} further confirm the contributions of our design choices.
When using slightly worse geometry from `w/ ‘w/o RGB Geo’' and `w/ ‘w/o NML Geo’', we can find rendering quality drops, which fits to our hypothesis that better deformed geometry will contribute to better rendering quality.
Notably, more input views obtain better performance, and `w/ 2 View' (front and back) is quality-wise closest to our four-view setting as with two views one can observe most of the body parts. 
\par \noindent\textbf{Runtime.}
In Fig.~\ref{fig:futureTeas}, we show runtime comparisons between our method and competing approaches on 1K and 4K resolutions. 
\textit{With a single RTX3090 GPU}, our method delivers superior rendering quality at a significantly faster runtime.
Notably, our method is three times faster than the prior state of the art HoloChar~\cite{shetty2024holoported}.
We refer to the supplementary material for detailed runtime analysis of individual modules.

\section{Discussion} \label{sec:conclusion}
\par \noindent\textbf{Limitations.} 
Although the proposed method achieves unprecedented rendering quality, speed, and robustness for sparse human capture and free-view rendering, there are remaining challenges opening up directions for future work.
First, our method can not maintain the rigidity of the fingers.
We can improve finger geometry by removing deformations on hands~\cite{shetty2024holoported} or estimating global hand translations.
Second, our method inherits the limitation of template-based methods~\cite{habermann2021real,remelli2022drivable,shetty2024holoported} in that it is hard to handle topological changes, like unzipping a jacket.
Third, the deformation model relies on scans or point-clouds for supervision, which require time-consuming per-frame reconstruction. 
Involving silhouette supervision~\cite{habermann2021real} or foundational depth priors~\cite{depthanything} may alleviate the need for this step.
\par \noindent\textbf{Conclusion.} 
Creating immersive and photorealistic renderings of real humans from sparse sensing data is of enormous importance as it has the potential to revolutionize the way of remote communication.
In this work, we take a step towards high-performance, robust, and photoreal human rendering from sparse inputs, which clearly outperforms the previous state of the art in all aspects.
At the core, our design deeply disentangles coarse geometric deformation and appearance prediction in a two stage process, which entirely operates in 2D texture space such that efficient parameter regression is possible at every stage.
For the future, we envision capturing more complex type of apparels, integrating motion estimation as part of the task, and reducing dense camera setups for training to solely sparse camera configurations.

\par \noindent\textbf{Acknowledgements}
This project was supported by the Saarbr\"ucken Research Center for Visual Computing, Interaction and AI, as well as the DFG's Research Training Group on “Neuroexplicit Models of Language, Vision, and Action” (GRK 2853/1).

{\small
\bibliographystyle{ieee_fullname}
\bibliography{mybib}
}

\clearpage
\appendix
\begin{figure}[tb]
	\includegraphics[width=0.98\linewidth] {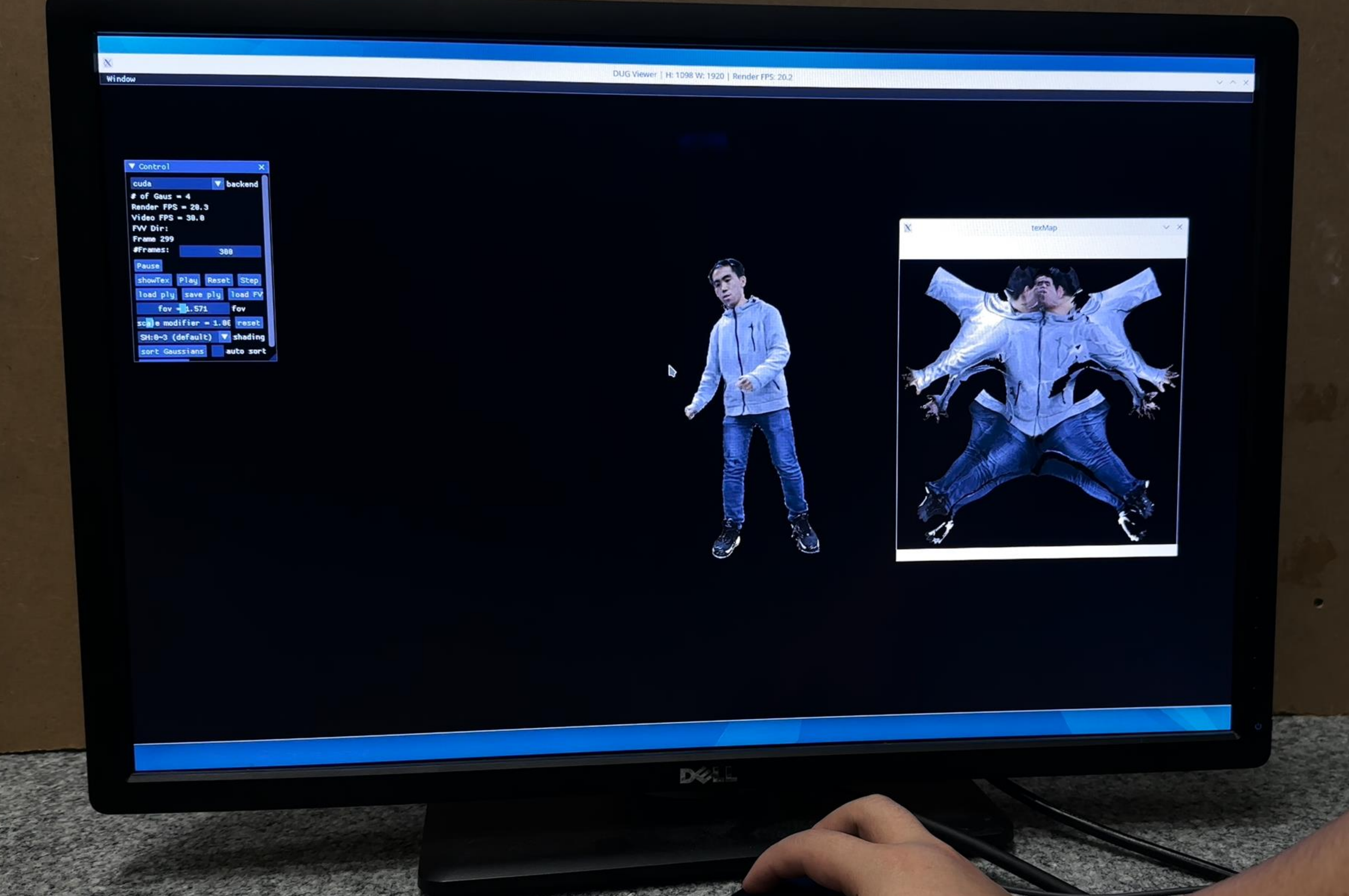}
	\caption
	{
        Given four-view image streams and body motions from the disk, our method generates photorealistic human renderings at real-time speed.
        }
	\label{fig:interactive}
\end{figure}

\section{Overview} 
We present more details and experimental results of DUT towards a more clear understanding and in-depth analysis.
We first show an interactive demo with our method (Sec.~\ref{sec:supp_interavtive}).
Then, we offer additional details about the visibility computation (Sec.~\ref{sec:supp_visibility}), loss function (Sec.~\ref{sec:supp_loss}), implementation details (Sec.~\ref{sec:supp_implementation}), runtime analysis (Sec.~\ref{sec:supp_runtime}), undeformed texture maps (Sec.~\ref{sec:supp_undeformed_texture}) and limitation discussions (Sec.~\ref{sec:supp_limitation}). 
We also provide additional experimental results about influence with respect to motion capture qualities (Sec.~\ref{sec:supp_mocap}), out-of-distribution motions (Sec.~\ref{sec:supp_ood}), motion sensitivity analysis (Sec.~\ref{sec:supp_sensitivity}), performance on a loose and long hair subject (Sec.~\ref{sec:supp_loose}), novel lighting (Sec.~\ref{sec:supp_lighting}).
Tab.~\ref{tab:notation} lists notations and symbols used in the main paper, and demonstrates their descriptions.
\section{Interactive Demo} \label{sec:supp_interavtive} 
To validate the potential of our method towards live telepresence, we built an interactive system to run our method in an end-to-end manner.
As illustrated in Fig.~\ref{fig:interactive} and also in the supplementary video, we load images and body motions from the disk of PC and perform live free-viewpoint rendering interactively controlled by the user.
The demo is performed on a PC with a single RTX3090 GPU and an Intel i9-10900X CPU.
\begin{figure*}[tb]
	\includegraphics[width=0.98\linewidth] {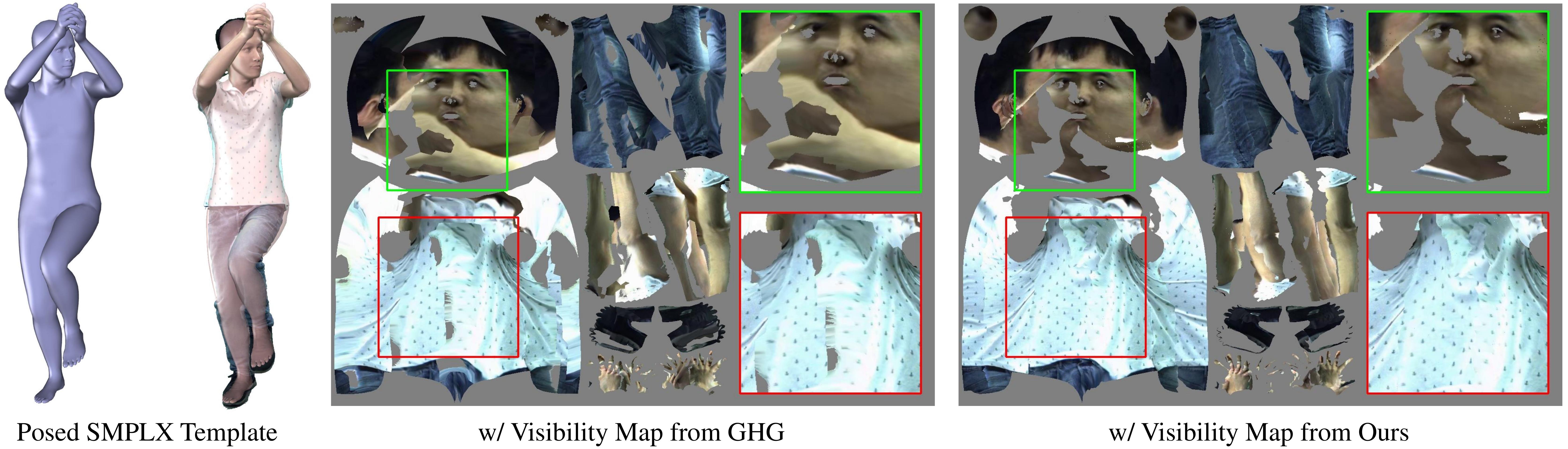}
	\caption
	{
        Compared to the visibility map computed by GHG~\cite{kwon2024generalizable}, the visibility map computed by our method is more robust and accurate, generating better unprojected texture map.
	   }
	\label{fig:unprojectionmap_with_ghg}
\end{figure*}

\section{Visibility Computation} \label{sec:supp_visibility} 
For methods~\cite{remelli2022drivable,shetty2024holoported,kwon2024generalizable} with texture unprojection, the computation of the visibility map is a crucial step, as the goal is to preserve as much information as possible while correctly unprojecting pixels into the texel space.
Here, we provide additional details about the visibility computation.
As presented in the Eq.~3, we use normal difference $\mathcal{T}_\mathrm{v}^{\mathrm{angle},i}$, depth difference $\mathcal{T}_\mathrm{v}^{\mathrm{depth},i}$, and segmentation masks $\mathcal{T}_\mathrm{v}^{\mathrm{mask},i}$ for the visibility computation criteria.
\par
The normal difference checks if an angle between the surface normal of the world space template and the inverse ray direction from the cameras are roughly parallel to each other. 
Similar to HoloChar~\cite{shetty2024holoported}, we render the normal texture maps $\mathcal{T}_\mathrm{N}^{i}$ and position texture maps $\mathcal{T}_\mathrm{P}^{i}$ of the world space template in texel space. 
The normal difference visibility is computed as:
\begin{align}
\mathcal{T}_\mathrm{ray}^{i} &=  \mathcal{T}_\mathrm{P}^{i} - \mathbf{o}^{i}\\
\mathcal{T}_\mathrm{v}^{\mathrm{angle},i} &= cos(-\mathcal{T}_\mathrm{ray}^{i}, \mathcal{T}_\mathrm{P}^{i}) > \delta
\end{align}
where $\mathbf{o}^{i}$ denotes the optical center of camera $i$ and $\mathcal{T}_\mathrm{ray}^{i}$ is the texture map storing ray directions. 
In practice, we set $\delta=0.17$ in all the experiments.
\par
We also perform depth difference verification to remove outlier points that are still on the same ray but on the backside of the template.
Following prior works~\cite{shetty2024holoported,remelli2022drivable}, we render depth texture maps $\mathcal{T}_\mathrm{D}^{i}$ and image coordinate texture maps $\mathcal{T}_\mathrm{xy}^{i}$ of the world space template in the texel space and also depth maps $\mathcal{I}_\mathrm{D}^{i}$ in the image space.
Then, we compute depth difference visibility as:
\begin{align}
\mathcal{T}_\mathrm{v}^{\mathrm{depth},i} &= \abs{ \pi_{uv}(\mathcal{T}_\mathrm{xy}^{i},\mathcal{T}_\mathrm{D}^{i})  - \mathcal{I}_\mathrm{D}^{i}}< \epsilon
\end{align}
In practice, we set $\epsilon=0.02$ in all the experiments.
\par
Apart from the geometry guidance, we can also utilize segmentation information to distinguish if points are in the foreground or background. 
Thus, the segmentation visibility can be computed as:
\begin{align}
\mathcal{T}_\mathrm{v}^{\mathrm{mask},i} &=  \pi_{uv}(\mathcal{T}_\mathrm{xy}^{i},\mathcal{I}_\mathrm{M}^{i})
\end{align}
where $\mathcal{I}_\mathrm{M}^{i}$ represents the segmentation of the image $\mathcal{I}^{i}$.
\par
Instead of geometry clues and segmentation clues, GHG~\cite{kwon2024generalizable} employs rasterized faces to determine visibility.
Specifically, they first rasterize the template's vertices onto image planes, and mark rasterized faces left on image planes as visible.
Next, they render the face visibility into the texel space to obtain the final visibility maps.
However, we found this routine is not stable, leading to incorrectly unprojected textures (Fig.~\ref{fig:unprojectionmap_with_ghg}).
Thus, to fully explore the upper bound of their method, we use our visibility computation method when re-implementing their method. 
\section{Loss Function} \label{sec:supp_loss} 
In this section, we discuss more details about loss functions and how they are computed.
\par \noindent \textbf{Chamfer Distance.} 
We use the Chamfer distance to evaluate the similarity between the vertices of posed deformed template $\mathbf{\hat{P}} = \mathbf{V}(\mathcal{M},\mathcal{D})$ and ground truth point-clouds $\mathbf{P}_\mathrm{GT}$:
\begin{align}
\mathcal{L}_\mathrm{Chamf} &= \frac{1}{N_\mathrm{V}} \sum_{p_{1} \in \mathbf{\hat{P}}} \min _{p_{2} \in \mathbf{P}_\mathrm{GT}}\|p_{1}-p_{2}\|_2^2 \nonumber \\ &+\frac{1}{\left|\mathbf{P}_\mathrm{GT}\right|} \sum_{p_{2} \in \mathbf{P}_\mathrm{GT}} \min _{p_{1} \in \mathbf{\hat{P}}}\|p_{1}-p_{2}\|_2^2
\end{align}
where $N_\mathrm{V}$ is the vertex number of the human template and $\left|\mathbf{P}_\mathrm{GT}\right|$ is the vertex number of $\mathbf{P}_\mathrm{GT}$.
\par \noindent \textbf{Laplacian Loss.} 
We apply a Laplacian loss on the posed and deformed template $\mathbf{V}(\mathcal{M},\mathcal{D})$ to ensure surface smoothness:
\begin{align}
\mathcal{L}_\mathrm{Lap} &= \frac{1}{N_\mathrm{V}} \sum\| \Delta(N_\mathrm{V},\mathbf{E}) \mathbf{V}(\mathcal{M},\mathcal{D})  \|_2^2,
\end{align}
where $\mathbf{E}$ represents the edges of the human template and $\Delta$ is the uniform Laplacian operator.
\par \noindent \textbf{Isometry Loss.} 
To prevent severe stretching of template edges, we constrain the edge length of the deformed template $\mathbf{T}_{\mathrm{D}}(\mathcal{D}, \mathbf{\bar{V}})$:
\begin{align}
\mathcal{L}_\mathrm{Iso} &= \frac{1}{N_\mathrm{V}} \sum_{i=1}^{N_\mathrm{V}}\frac{1}{\left|\mathcal{E}_{i}\right|}\sum_{ e_{i,j} \in \mathcal{E}_{i}}\| \mathbf{\bar{V}}(e_{i,j}) - \mathbf{T}_{\mathrm{D}}(\mathcal{D}, \mathbf{\bar{V}})(e_{i,j})  \|_2^2,
\end{align}
where $\mathcal{E}_{i}$ is the edge set of vertex ${i}$ and $\left|\mathcal{E}_{i}\right|$ is the edge number of $\mathcal{E}_{i}$. 
$\mathbf{\bar{V}}(e_{i,j})$ represents indexing the vertices of edge $e_{i,j}$ from $\mathbf{\bar{V}}$.
\par \noindent \textbf{Normal Consistency Loss.} 
A normal consistency loss is used to improve the consistency between the normal of a vertex and its nearby vertices:
\begin{align}
\mathcal{L}_\mathrm{Nc} &=  \frac{1}{N_\mathrm{V}} \sum_{i=1}^{N_\mathrm{V}}\frac{1}{\left|\mathcal{N}_{i}\right|}\sum_{ v_{i,j} \in \mathcal{N}_{i}}\| 1-cos( \mathbf{N}_{\mathrm{D}}(v_{i}), \mathbf{N}_{\mathrm{D}}(v_{i,j}))  \|_2^2,
\end{align}
where $\mathcal{N}_{i}$ is the set of nearby vertices of vertex ${i}$, $\left|\mathcal{N}_{i}\right|$ is the number of $\mathcal{N}_{i}$, and $\mathbf{N}_{\mathrm{D}}(v_{i})$ denotes the normal of vertex $i$ on the deformed template $\mathbf{T}_{\mathrm{D}}(\mathcal{D}, \mathbf{\bar{V}})$.
\par \noindent \textbf{L1 Loss.} 
We compute the L1 loss between the rendered image $\mathcal{\hat{I}}'$ and the ground truth image $\mathcal{I}$ as:
\begin{align}
\mathcal{L}_\mathrm{L1} &=  \frac{1}{\left| \mathcal{I} \right|} \left| \mathcal{\hat{I}}' -   \mathcal{I} \right|.
\end{align}
\par \noindent \textbf{SSIM Loss.} 
The SSIM loss is mainly used to maintain the structure similarity between the rendered image $\mathcal{\hat{I}}'$ and the ground truth image $\mathcal{I}$:
\begin{align}
\mathcal{L}_\mathrm{SSIM} &=  1 - \mathbf{SSIM} (\mathcal{\hat{I}}',   \mathcal{I} ).
\end{align}
\par \noindent \textbf{IDMRF Loss.} 
Following prior works~\cite{feng2022capturing,xiang2023drivable, chen2024egoavatar}, we additionally use the IDMRF Loss~\cite{wang2018image} for the perceptual regularization and encourage high-frequency details.
\begin{table*}[h]
\large
\setlength\tabcolsep{2pt}
    \centering
    \normalsize
    \scalebox{0.95}{
    \begin{tabular}{|c|c|c|c|c|c|c|c|c|c|}
    \hline \multirow{3}{*}{ GPU } & \multirow{3}{*}{ Rnd Res } & \multirow{3}{*}{ Tex Res } & \multicolumn{3}{c|}{ Stage I } & \multicolumn{3}{c|}{ Stage II } & \multirow{3}{*}{ FPS }\\
    \cline { 4 - 9 } & & &  \begin{tabular}{c} 
    Forward \\
    Kinematic
    \end{tabular} & \begin{tabular}{c} 
    Obtaining First \\
    Unprojected Map
    \end{tabular} & \begin{tabular}{c} 
    GeoNet \\
    Inference
    \end{tabular} & \begin{tabular}{c} 
    Obtaining Second \\
    Unprojected Map
    \end{tabular} & \begin{tabular}{c} 
    GauNet \\
    Inference
    \end{tabular} & Rendering & \\
    \hline 3090 & 1 K & 256 & 3.8723 & 7.6027 & 11.4430 & 14.6424 & 21.3143 & 23.5674 & 42.43 \\
    \hline 3090 & 4 K & 256 & 3.8751 & 7.6097 & 11.5076 & 14.7454 & 21.4354 & 23.7146 & 42.17 \\
    \hline 3090 & 1 K & 512 & 3.7012 & 7.3747 & 11.2363 & 14.2890 & 35.0456 & 37.1450 & 26.92 \\
    \hline 3090 & 4 K & 512 & 3.7457 & 7.4154 & 11.2667 & 14.3009 & 35.0166 & 37.0996 & 26.95 \\
    \hline H100 & 1 K & 256 & 3.8130 & 6.9313 & 10.7382 & 13.3484 & 16.5124 & 18.7570 & 53.31 \\
    \hline H100 & 4 K & 256 & 3.8489 & 6.9685 & 10.7392 & 13.3512 & 16.4705 & 18.7303 & 53.39 \\
    \hline H100 & 1 K & 512 & 3.7181 & 6.9793 & 10.6773 & 13.3481 & 19.5077 & 21.7711 & 45.93 \\
    \hline H100 & 4 K & 512 & 3.6698 & 6.8365 & 10.4528 & 13.0562 & 19.1951 & 21.3978 & 46.73 \\
    \hline
    \end{tabular}
    }
    \caption{
    \textbf{Quantitative Ablation.} Here, we demonstrate detailed runtime ablation of our methods with different texture resolutions (Tex Res), rendering resolutions (Rnd Res) and GPUs.
    All the time in the table is the accumulated time from the beginning and  their units are milliseconds.
\label{tab:time} 
    }
\end{table*}

\section{Implementation Details} \label{sec:supp_implementation} 
\par \noindent \textbf{Motion Capture and Template Tracking.}
We use a marker-less motion capture approach~\cite{stoll2011fast, captury} to recover the body motions for our method and HoloChar~\cite{shetty2024holoported}, which takes images from 34 cameras as input.
DVA~\cite{remelli2022drivable} and GHG~\cite{kwon2024generalizable} require SMPLX~\cite{SMPL-X:2019} as the human template. 
To eliminate the influence of motion capture accuracy, we first transform motions from our format to SMPLX format and refine the shape parameters and body parameters with ground truth point-clouds from NeuS2~\cite{neus2}.
Notably, our method does not need ground truth point-clouds for motion refinement and can still work with sparse-view motion capture results as inputs (Sec.~\ref{sec:supp_mocap}). 
\par \noindent \textbf{Training and Evaluation.}
The training views, condition views, and evaluation views do not overlap, expect for S2618.
For S2618, the condition views are also used as supervisions during training due to the limited number of available cameras in this dataset.
The training sequences and evaluation sequences do not overlap, but they share similar types of motion.
For the newly collected subject, we also include novel action types that are completely out of the distribution of action types in the training set (Tab.~\ref{tab:action}).
\par \noindent \textbf{Our Method.} 
We implement all the modules of our method in Pytorch~\cite{paszke2017automatic}. 
To accelerate the computation of the texture unprojection, we use nvdiffrast~\cite{Laine2020diffrast} for parallel rendering and implement forward kinematics~\cite{Pang_2024_CVPR} and camera projection with the extension of Pytorch.
The visualizer of our interactive demo is built upon 3DGStream~\cite{sun20243dgstream}.
\begin{table*}[h]
\large
\setlength\tabcolsep{2pt}
    \centering
    \normalsize
    \scalebox{0.95}{
    \begin{tabular}{|c|c|c|c|}
        \hline
        
        \multicolumn{4}{|c|}{\textbf{Training \& Testing Action}}\\
        \hline
        Jogging & Walking & Looking & Picking Up \\
        \hline
         Talking & Waving & Celebrating & Jumping \\ 
        \hline
        Baseball Throwing  & Baseball Swing  & Boxing & Goalkeeping \\
        \hline
        Penalty Kick & Golf & Archery & Weight Lifting \\ 
        \hline
         Squating & Jumping Jack & Playing Instrument & Dancing  \\ 
        \hline
         Opening and Pushing Door  & Playing Table Tennis & Playing Badminton & Giving Presentation \\
        \hline
        Drinking & Using Phone & Petting Animal & Playing Hockey\\ 
        \hline
         Playing Tug of War & Juggle Balls  & Playing Hula Hoop & Bowling  \\ 
        \hline
         Playing Volleyball & Wrestling & Stretching & Mopping Floor  \\
        \hline
        Digging & Typing & Cooking & Using Spray \\
        \hline
        Applying Makeup &  & & \\
        \hline
        \multicolumn{4}{|c|}{\textbf{Out of Distribution Action}}\\
        \hline
        Shooting & Surrender & Fishing & Standing Long Jump\\
        \hline
         Single-leg Hop & Frog Jump & Crawling & Rolling on the Ground\\
        \hline
         Sleeping & Ultraman &  & \\ 
        \hline
    \end{tabular}
    }
    \caption{
    The action types of body motions in the training split, testing split, and out-of-distribution split.
\label{tab:action} 
    }
\end{table*}

\begin{table}[t]
  \centering
      \small
          \scalebox{0.9}{
    \begin{tabular}{|c|c|c|c|c|c|c|}
    \hline
    \textbf{Methods} & Motion & \textbf{Tex Res} &  \textbf{PSNR} $\uparrow$ & \textbf{SSIM} $\uparrow$ & \textbf{LPIPS} $\downarrow$  \\
    \hline
     Ours-Large &  Sparse & 512 & \silve{30.0309}	& \gold{0.8722} &	\gold{0.1322}  \\
    \hline
     Ours-Large &  Dense & 512 & \gold{30.0311}	& \gold{0.8722} &	\gold{0.1322}  \\
    \hline
    \end{tabular}
    }
  \caption{
  \textbf{Quantitative Ablation.}
  We evaluate the influence of the motion capture quality.
  Our method still produces reasonable results with sparse-view captured motions.
  }
  \label{tab:ablation_motion}
  \normalsize
\end{table}

\begin{figure*}[tb]
        \centering
	\includegraphics[width=\linewidth] {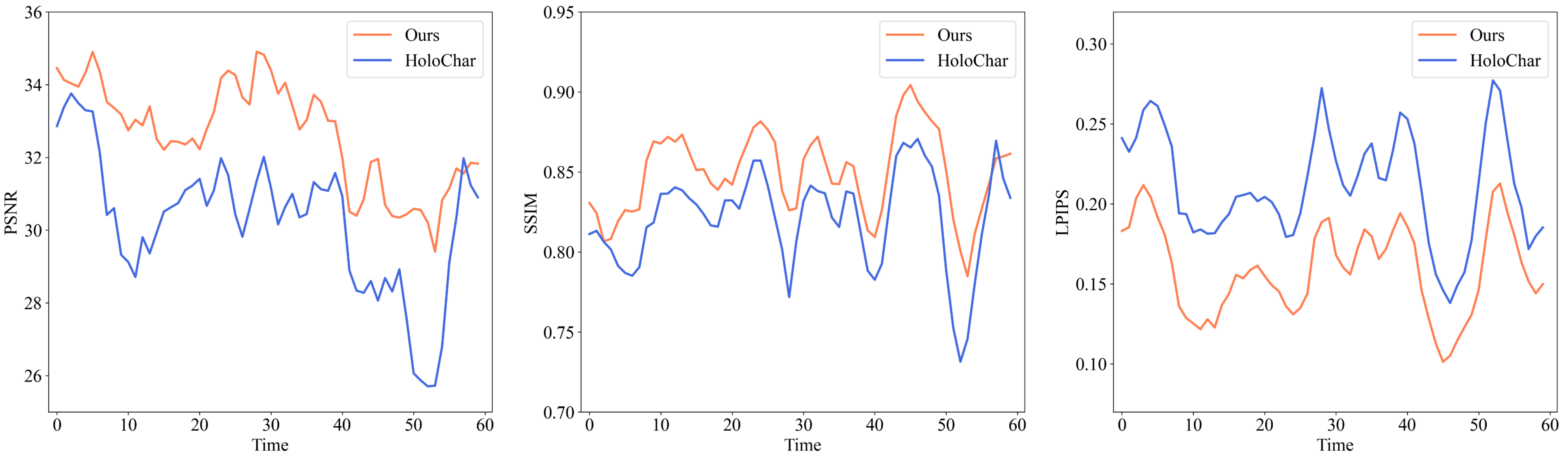}
	\caption
	{
        \textbf{Quantitative Comparison.}
        We quantitatively compare the rendering results of our method and HoloChar~\cite{shetty2024holoported} on out-of-distribution human motions.
        Our method produces consistently better rendering results.
	}
	\label{fig:ood_curve}
\end{figure*}

\par \noindent \textbf{GHG.} 
GHG~\cite{kwon2024generalizable} is a method for generalizable human rendering.
We use the offical code and the official checkpoint of their inpainting network. 
With our tracked SMPLX and refined visibility map, we freeze the weights of the inpainting network and train the Gaussian regressor for each subject, respectively.
\par \noindent \textbf{HoloChar.} 
For each subject, we first train the geometry module~\cite{habermann2021real} of HoloChar~\cite{shetty2024holoported} with, both, point-clouds and distance transformation images~\cite{borgefors1986distance}, and then use the official code of HoloChar~\cite{shetty2024holoported} to train the texture network and the super resolution network.
\par \noindent \textbf{DVA.} 
With the official code, we train DVA~\cite{remelli2022drivable} with our tracked SMPLX and same train and test splits.
\par \noindent \textbf{ENeRF.} 
Though ENeRF~\cite{lin2022efficient} is not specifically designed for human rendering, their method has some generalization ability.
For each subject, we provide the same conditioning inputs (four-view images) as our method and finetune the provided official checkpoint. 
\section{Runtime Analysis} \label{sec:supp_runtime} 
In this section, we discuss the detailed runtime performance of each method. 
Since all methods differ in network structures and modeling strategies, we design a runtime evaluation protocol for a fair comparison.
More specifically, we neglect the data loading time and assume an ideal situation that a method receives data, processes data, and outputs rendering results.
For each method and component, we calculate the average runtime over 100 frames, and repeat this process three times to obtain the final average runtime.
\par \noindent \textbf{ENeRF.}
After receiving the conditioning images and novel camera parameters, ENeRF~\cite{lin2022efficient} can be split into two stages: ray sampling and model inference (including rendering).
For 1K resolution, the runtime of ray sampling is 7.84 ms and the runtime of model inference is 30.83 ms, the total runtime is 38.68 ms and 25.85 FPS.
For 4K resolution, the runtime of ray sampling is 158.68 ms and the runtime of model inference is 351.03 ms, the total runtime is 509.72 ms and 1.96 FPS.
\par \noindent \textbf{DVA.}
We can also divide DVA~\cite{remelli2022drivable} into two stages: model inference and rendering.
For 1K resolution, the runtime of model inference is 33.90 ms and the runtime of rendering is 7.53 ms, the total runtime is 41.43 ms and 24.13 FPS.
For 4K resolution, the runtime of model inference is 178.45 ms and the runtime of rendering is 7.62 ms, the total runtime is 186.07 ms and 5.37 FPS.
\par \noindent \textbf{HoloChar.}
HoloChar~\cite{shetty2024holoported} includes three stages, geometry inference, texture unprojection, and texture inference.
Their 1K rendering and 4K rendering are performed together, thus we only report the runtime for 4K.
The runtime of geometry inference is 9.34 ms, the runtime of texture unprojection is 28.35 ms and the runtime of texture inference is 35.75 ms.
The total runtime is 73.46 ms and 13.61 FPS.
Notably, we perform all the modules on a workstation with a single GPU and obtain slightly better inference speed, compared to the runtime evaluation in the original paper.
\begin{figure*}[tb]
        \centering
	\includegraphics[width=0.95\linewidth] {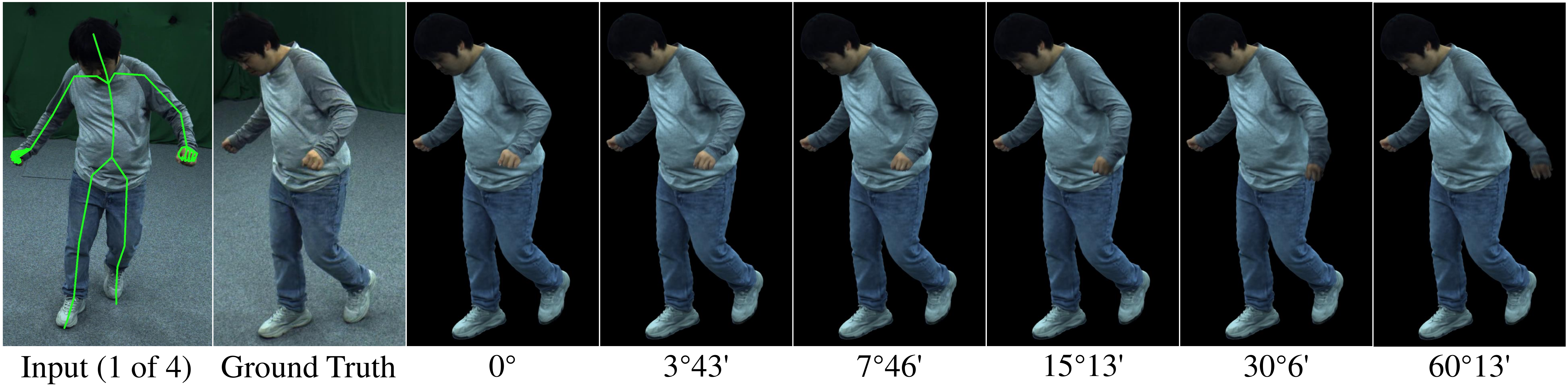}
	\caption
	{
        \textbf{Motion Sensitivity Analysis to Increasing Errors}.
        Our method still outputs reasonable results to different level of motion capture errors at inference.
	}
	\label{fig:motion_sensitivity}
\end{figure*}

\begin{figure*}[tb]
        \centering
	\includegraphics[width=0.95\linewidth] {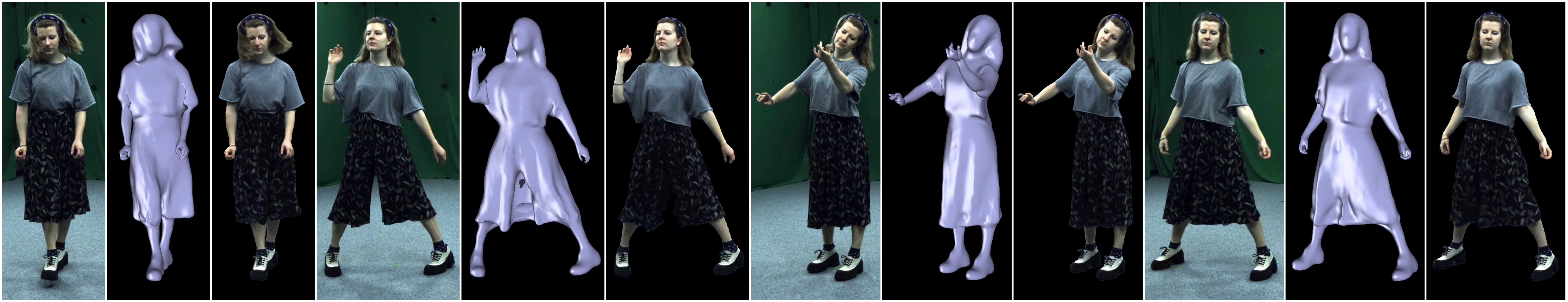}
	\caption
	{
        \textbf{Qualitative results.}
        We perform our method on a loose and long-hair subject, it manages to capture the coarse deformations of hair and dress and produces faithful rendering results.
	}
	\label{fig:loose_subject}
\end{figure*}

\begin{figure*}[tb]
        \centering
	\includegraphics[width=0.98\linewidth] {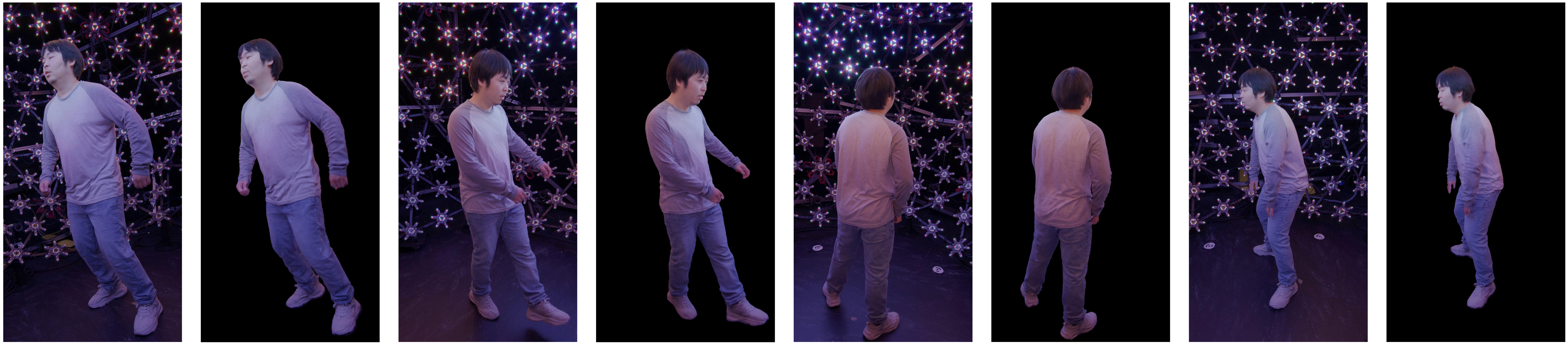}
	\caption
	{
        \textbf{Qualitative results}.
        Results with finetuned model under novel lighting. After finetuning, DUT still runs in feed-forward manner.
	}
	\label{fig:novel_lighting}
\end{figure*}

\begin{figure*}[tb]
        \centering
	\includegraphics[width=0.98\linewidth] {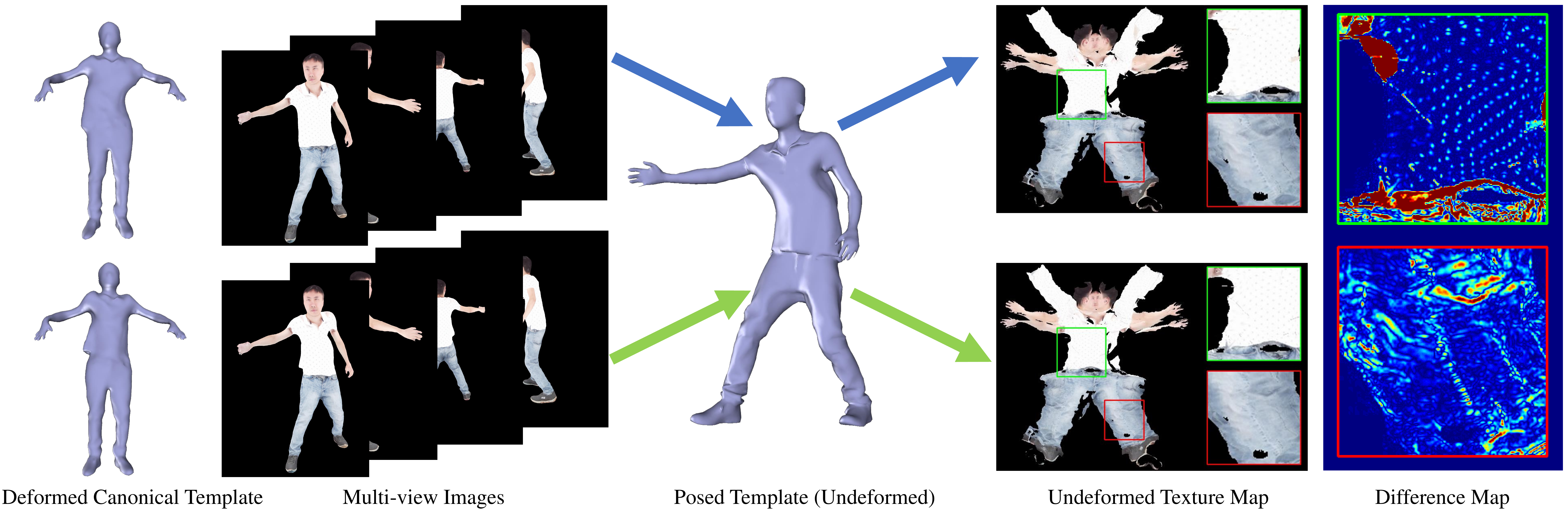}
	\caption
	{
        \textbf{Illustration of Undeformed Texture Map}.
        The distortions of undeformed (first) texture maps are directly related with deformations on the canonical template. 
	}
	\label{fig:undeformed}
\end{figure*}

\begin{figure*}[!]
        \centering
	\includegraphics[width=0.95\linewidth]         
        {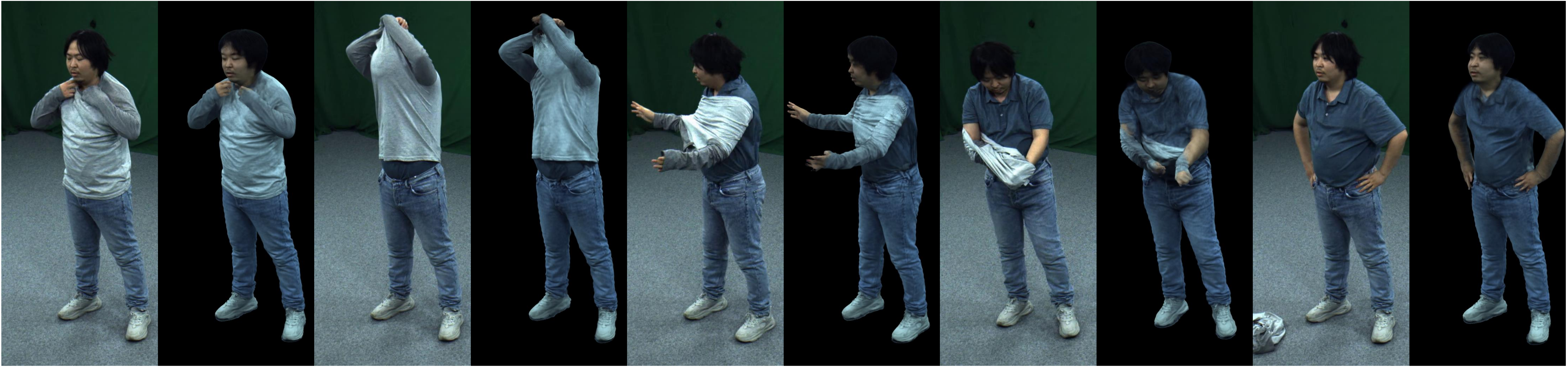}
	\caption
	{
        \textbf{Topology Change}.
        Results of taking off cloth.
	}
	\label{fig:topology_change}
\end{figure*}

\par \noindent \textbf{GHG.}
GHG~\cite{kwon2024generalizable} has four stages, position map rendering, visibility map rendering, network inference, and rendering.
Though, in the experiments, we use our visibility computation to obtain better performance of GHG (Sec.~\ref{sec:supp_visibility}).
Here, we only report the runtime performance of their original implementation.
The runtime of position map rendering is 922.10 ms and the runtime of visibility map rendering is 2754.64 ms.
For 1K resolution, the runtime of model inference is 246.83 ms and the runtime of rendering is 214.37 ms.
The total runtime is 4137.96 ms and 0.2416 FPS.
For 4K resolution, the runtime of model inference is 247.14 ms and the runtime of rendering is 213.87 ms.
The total runtime is 4137.77 ms and 0.24 FPS.
\par \noindent \textbf{Ours.}
Our method contains two main stages: geometry stage and appearance stage. 
Towards a more comprehensive runtime analysis, we will split each stage into fine-grained components and report the accumulated time for each component.
As illustrated in Tab.~\ref{tab:time}, our method finishes all the operations within 38 ms.
The rendering resolution does not affect the speed of our method.
Besides, we found our method has the potential for improvement when running it on more powerful GPUs, i.e. an NVidia H100 graphics card.

\section{Ablation on Mocap Quality} \label{sec:supp_mocap} 
In Tab.~\ref{tab:ablation_motion}, we evaluate how the motion capture quality affects the performance of our method on S3.
We use the same motion capture method~\cite{captury,stoll2011fast} as before, but only provide 4 input views.
There is a small performance drop when replacing dense motion capture with sparse motion capture on the PSNR metric.
We believe this is due to the fact that current motion capture methods could produce slightly worse results for the sparse-view setting, while our method can still compensate for such errors to some extent. 
\section{Additional Comparisons on OOD Motions} \label{sec:supp_ood} 
In Fig.~\ref{fig:ood_curve}, we show additional quantitative comparisons between our method and HoloChar~\cite{shetty2024holoported} on the sequences with out-of-distribution motions.
Our method produces consistently better results, and also lower standard deviations.
In terms of PSNR, our method's standard deviation is $\textbf{1.41}$, while HoloChar's results have a standard deviation of $1.89$, which reveals the robustness of our method on OOD motions.
\section{Sensitivity Analysis to Motion Errors} \label{sec:supp_sensitivity} 
We show a sensitivity analysis by manually adding linear motion capture errors to left elbow from 0$^{\circ}$ to $60^{\circ}13^{'}$
in Fig.~\ref{fig:motion_sensitivity}.
Surprisingly, DUT produces good results within $7^{\circ}46^{'}$ error (0$^{\circ}$, $3^{\circ}43^{'}$, $7^{\circ}46^{'}$) and reasonable results even with $60^{\circ}13^{'}$ error.

\section{Performance on Loose Clothing and Long Hair Subject} \label{sec:supp_loose}
In Fig.~\ref{fig:loose_subject}, we show a subject with dynamic long-hair wearing a dress.
Our geometry module recovers the coarse geometry of hair dynamics and dress.
After which, our rendering module faithfully generates high fidelity renderings.

\section{Results under Novel Lighting Environment} \label{sec:supp_lighting}
Towards realistic application scenarios, once our method is trained, we may run it with novel cameras under novel lighting environments.
We can either perform color augmentation to the input views during training~\cite{xiang2023drivable} or finetune the model on the sparse input views captured under the new illumination.
We provide a preliminary result for this in Fig.~\ref{fig:novel_lighting} where we finetuned the model on the input views and \textbf{isolated frames (not used in testing)} in a new lighting condition.
After finetuning, DUT still runs in feed-forward manner.
Our method can be easily adopted to the novel lighting and produce reasonable rendering results.

\section{Additional Discussions about Undeformed (First) Texture Map} \label{sec:supp_undeformed_texture}
Our GeoNet $\Phi_{\mathrm{Geo}}$ estimates template deformations from undeformed texture map $\mathcal{T}_\mathrm{c, 1st}$ and non-root normal map $\mathcal{T}_\mathrm{\bar{N}}$.
Similar to the normalized skeletal motion in DDC~\cite{habermann2021real}, $\mathcal{T}_\mathrm{\bar{N}}$ provides the pose information.
However, dynamic geometries of a moving human body are not completely determined by the skeletal pose at that moment, it leads to one-to-many mapping issue~\cite{liu2021neural}.
Our undeformed texture map $\mathcal{T}_\mathrm{c, 1st}$ offers additional information about the degree of deformations. 
As shown in Fig.~\ref{fig:undeformed}, under the same body pose, the degree of deformations can be reflected by the distortions of $\mathcal{T}_\mathrm{c, 1st}$.

\section{More Discussions about Limitations} \label{sec:supp_limitation}

\par \noindent\textbf{Color Fluctuation.} 
Though simple and efficient, our method suffers from a certain extent color fluctuation of some frames as shown in the video.
We found that this could be attributed to the predicted Gaussians are trying to overfit the uneven lightings of studio and shadows on the body, which are challenging for such simple representation.
Integrating ray tracing or physically based rendering may reduce the color fluctuation.

\par \noindent\textbf{Topology Change.} 
In Fig.~\ref{fig:topology_change}, we show results on a sequence where the subject is taking clothes off.
While results look reasonable, still the quality degrades.
Though a fixed template contributes a lot, it will be an interesting direction to investigate how to introduce dynamic template update into such task, especially with only RGB inputs.
\begin{table*}[h]
\setlength\tabcolsep{2pt}
    \centering
    \scalebox{0.95}{
    \begin{tabular}{|c|c|}
        \hline
        \textbf{Symbol} & \textbf{Description} \\
        \hline
        $\mathcal{D}$ & Deformation parameters \\
        $\mathcal{M}$ & Body motion parameters \\
        $\mathbf{\bar{V}}$ &  Vertices of human template mesh in the canonical body pose \\
        $N_\mathrm{V}$ & Number of vertices on human template mesh \\
        $\mathbf{T}_{\mathrm{LBS}}(\cdot)$ & LBS transformation function\\
        $\mathbf{T}_{\mathrm{D}}(\cdot)$ & Mesh deformation function\\
        $\mathbf{H}(\cdot)$ & A function that converts spherical harmonics coefficients $\textbf{h}$ into RGB colors $\textbf{s}$ \\
        $x$ & 3D positions in the Euclidean space \\
        $\mathbf{R}$  & 3D rotation matrix \\
        $\mathbf{S}$  & 3D scale matrix \\
        $\mathbf{J}$  & Jacobian of the affine approximation of the projective transformation \\
        $\mathbf{W}$  & 3D view matrix \\
        $\textbf{G}$ & A set of Gaussian parameters, including $\textbf{p}$, $\textbf{r}$, $\textbf{h}$, $\textbf{s}$, $\boldsymbol{\alpha}$ \\
        $\textbf{p}$ & Positions of 3D Gaussians in world space \\
        $\textbf{r}$ & Rotations of 3D Gaussians \\
        $\textbf{h}$ & Spherical harmonics of 3D Gaussians \\
        $\textbf{s}$ & Scales of 3D Gaussians \\
        $\boldsymbol{\alpha}$ & Density values of 3D Gaussians \\
        $\boldsymbol{\alpha}^{'}$ & Density values of 3D Gaussians in 2D\\
        $\textbf{c}$ & Color values of 3D Gaussians \\          
        $\mathcal{T}_\mathrm{c}$ & Unprojected texture map in texel space \\
        $\mathcal{T}_\mathrm{v}$ & Visibility map in texel space\\
        $\mathcal{T}_\mathrm{v}^{\mathrm{angle}}$ & Visibility map computed by normal difference in texel space\\
        $\mathcal{T}_\mathrm{v}^{\mathrm{depth}}$ & Visibility map computed by depth difference in texel space\\
        $\mathcal{T}_\mathrm{v}^{\mathrm{mask}}$ & Visibility map computed by segmentation mask in texel space\\
        $\Phi_{\mathrm{Geo}}$ & Geometry network that estimates deformations of human body template in texel space \\
        $\Phi_{\mathrm{Gau}}$ & Gaussian network that estimates Gaussian parameters in texel space \\
        $\mathcal{T}_\mathrm{\bar{N}}$ & Normal map of posed template but without root rotation in texel space \\
        $\mathcal{T}_\mathrm{c, 1st}$ & First unprojected texture map in texel space \\
        $\mathcal{T}_\mathrm{c, 2nd}$ & Second unprojected texture map in texel space \\
        $\boldsymbol{\mathcal{G}}$ & A set of modified Gaussian parameters, including $\textbf{d}$, $\textbf{r}$, $\textbf{h}$, $\textbf{s}$, $\boldsymbol{\alpha}$ \\
        $\mathcal{M}_\mathcal{G}$ & Mask map of valid Gaussians in texel space \\
        $R(\cdot)$ & Gaussian renderer \\
        $\mathcal{T}_\mathrm{LBS}$ & LBS transformations map in texel space \\
        $\mathcal{T}_{\mathbf{T}_{\mathrm{D}}(\mathcal{D}, \mathbf{\bar{V}})}$ & Deformed base geometry map in texel space \\
        $\mathcal{T}_{\mathcal{G}}$ &  Modified Gaussian map $\boldsymbol{\mathcal{G}}$ in texel space\\
        $\mathcal{T}_{{\boldsymbol{\mathcal{G}}},\mathrm{\chi}}$ & Feature map  $\mathrm{\chi}$ of $\mathcal{T}_{\mathcal{G}}$ in texel space, $\mathrm{\chi} \in \{\textbf{d},\textbf{h},\textbf{s},\textbf{r},\boldsymbol{\alpha} \}$\\
        $\pi_{uv}(\cdot)$ & The function that indexes the features in the texel space. \\
        $\mathcal{\hat{I}}'$ &  Rendered image with Gaussian scale refinement\\
        $\mathcal{T}_{\mathrm{s}'}$ & Refining scale map in texel space\\
        $\mathcal{L}_\mathrm{Chamf}$ & Chamfer distance  \\
        $\mathcal{L}_\mathrm{Lap}$ & Laplacian loss  \\
        $\mathcal{L}_\mathrm{Iso}$ & Isometry loss  \\
        $\mathcal{L}_\mathrm{Nc}$ & Normal consistency loss  \\
        $\mathcal{L}_\mathrm{L1}$ & L1 loss \\
        $\mathcal{L}_\mathrm{SSIM}$ & SSIM loss  \\
        $\mathcal{L}_\mathrm{IDMRF}$ & IDMRF loss  \\
        $\mathcal{L}_\mathrm{Reg}$ & Geometric regularization loss  \\
        \hline
    \end{tabular}
    }
    \caption{
    Notations and symbols.
    }
    \label{tab:notation}
\end{table*}

\end{document}